\documentclass[letterpaper,journal]{IEEEtran}
\usepackage{amsmath,amsfonts}
\usepackage{algorithmic}
\usepackage{algorithm}
\usepackage{array}
\usepackage[caption=false,font=normalsize,labelfont=sf,textfont=sf]{subfig}
\usepackage{textcomp}
\usepackage{stfloats}
\usepackage{url}
\usepackage{verbatim}
\usepackage{graphicx}
\usepackage{cite}
\usepackage{xcolor}
\usepackage{multirow}
\usepackage{booktabs}
\usepackage{makecell}
\usepackage{pifont}
\usepackage{amssymb}
\usepackage{svg}
\usepackage{hyperref}
\hypersetup{
    colorlinks=true,      
    linkcolor=black,        
    citecolor=black,       
    urlcolor=magenta,     
    filecolor=black        
}

\begin{document}

\title{PolarVSR: A Unified Framework and Benchmark for Continuous Space-Time Polarization Video Reconstruction}

\author{Chenggong~Li,
        Yidong~Luo,
        Junchao~Zhang$^{\dagger}$,~\IEEEmembership{Member,~IEEE},
        Boxin~Shi,~\IEEEmembership{Senior~Member,~IEEE},
        and Degui~Yang$^{\dagger}$%
\thanks{Chenggong Li, Junchao Zhang, and Degui Yang are with the School of Automation, Central South University, Changsha, China, and also with the Hunan Provincial Key Laboratory of Optic-Electronic Intelligent Measurement and Control, Changsha, China (e-mail: 244603040@csu.edu.cn; junchaozhang@csu.edu.cn; degui.yang@csu.edu.cn).}%
\thanks{Yidong Luo is with Zhejiang University, Hangzhou, China, and also with the School of Engineering, Westlake University, Hangzhou, China (e-mail: luoyidong@westlake.edu.cn).}%
\thanks{Boxin Shi is with the State Key Laboratory of Multimedia Information Processing, School of Computer Science, Peking University, Beijing, China, and also with the National Engineering Research Center of Visual Technology, School of Computer Science, Peking University, Beijing, China (e-mail: shiboxin@pku.edu.cn).}%
\thanks{$^{\dagger}$Junchao Zhang and Degui Yang are co-corresponding authors.}}

\markboth{arXiv preprint}%
{Li \MakeLowercase{\textit{et al.}}: PolarVSR: A Unified Framework and Benchmark for Continuous Space-Time Polarization Video Reconstruction}


\maketitle

\begin{abstract}
Polarimetric imaging captures surface polarization characteristics, such as the Degree of Linear Polarization (DoLP) and the
Angle of Polarization (AoP). In mainstream Division-of-Focal-Plane (DoFP) color polarization imaging, recovering polarization parameters
from captured mosaic arrays remains a challenging inverse problem. Existing DoFP cameras also face hardware bottlenecks and often
cannot support high-frame-rate acquisition, limiting polarimetric imaging in dynamic video tasks. These limitations motivate
joint spatial and temporal enhancement. To this end, we propose the first space-time polarization video reconstruction architecture.
The method jointly models polarization directions in space and time and uses a polarization-aware implicit neural representation for
continuous, high-fidelity upsampling. By analyzing temporal variations in polarization parameters, we further introduce a flow-guided
polarization variation loss to supervise polarization dynamics. We also establish the first large-scale color DoFP polarization video
benchmark to support this research direction. Extensive experiments on this benchmark demonstrate the effectiveness of the method.
The source code and benchmark will be released at \href{https://jjgnb.github.io/PolarVSR.github.io/}{this project page}.
\end{abstract}


\section{Introduction}
\IEEEPARstart{P}{olarization} of light encodes intrinsic surface properties, including reflectivity, roughness, material composition, and geometry. 
Unlike conventional RGB intensity, which mainly records appearance and illumination, polarization provides complementary light-surface interaction cues for vision tasks such as object detection~\cite{luo2025cpifuse}, semantic segmentation~\cite{liu2025sharecmp}, reflection removal~\cite{yao2025polarfree,lyu2022physics,lei2020polarized}, and surface normal estimation~\cite{deschaintre2021deep,lyu2023shape,lei2022shape}. 
Most existing studies focus on still polarization images, whereas many real-world applications require continuous perception of dynamic scenes. 
Polarization videos capture the temporal evolution of polarimetric cues caused by camera motion, object motion,
viewpoint changes, and time-varying reflections, making them desirable for video-level vision systems such as
autonomous driving \cite{chen2025rethinking}, robotic perception \cite{ponimatkin20256d},
motion understanding \cite{hong2025motionbench}, and 3D reconstruction \cite{liu2025slam3r}.

Mainstream Division-of-Focal-Plane (DoFP) color polarization cameras (e.g., FLIR BFS-U3-51S5PC-C) \cite{rebhan2019principle} capture
intensities from four polarization directions in one shot. These measurements are converted into Stokes vectors \cite{collett2005field}
to compute key polarimetric parameters, such as the Degree of Linear Polarization (DoLP) and the Angle of Polarization (AoP), for downstream
applications. Despite the benefits of polarization imaging, DoFP hardware imposes two major constraints. Spatially, mosaic sampling discards
substantial scene information, yielding a much lower effective resolution than standard color cameras. Temporally, current sensors have limited
frame rates, restricting DoFP cameras in video-level tasks. As video-level vision systems emerge, conventional per-frame restoration cannot meet
modern input demands, making unified space-time enhancement important for deployment.

\begin{figure*}[!t]
	\begin{center}
		\includegraphics[width=\linewidth]{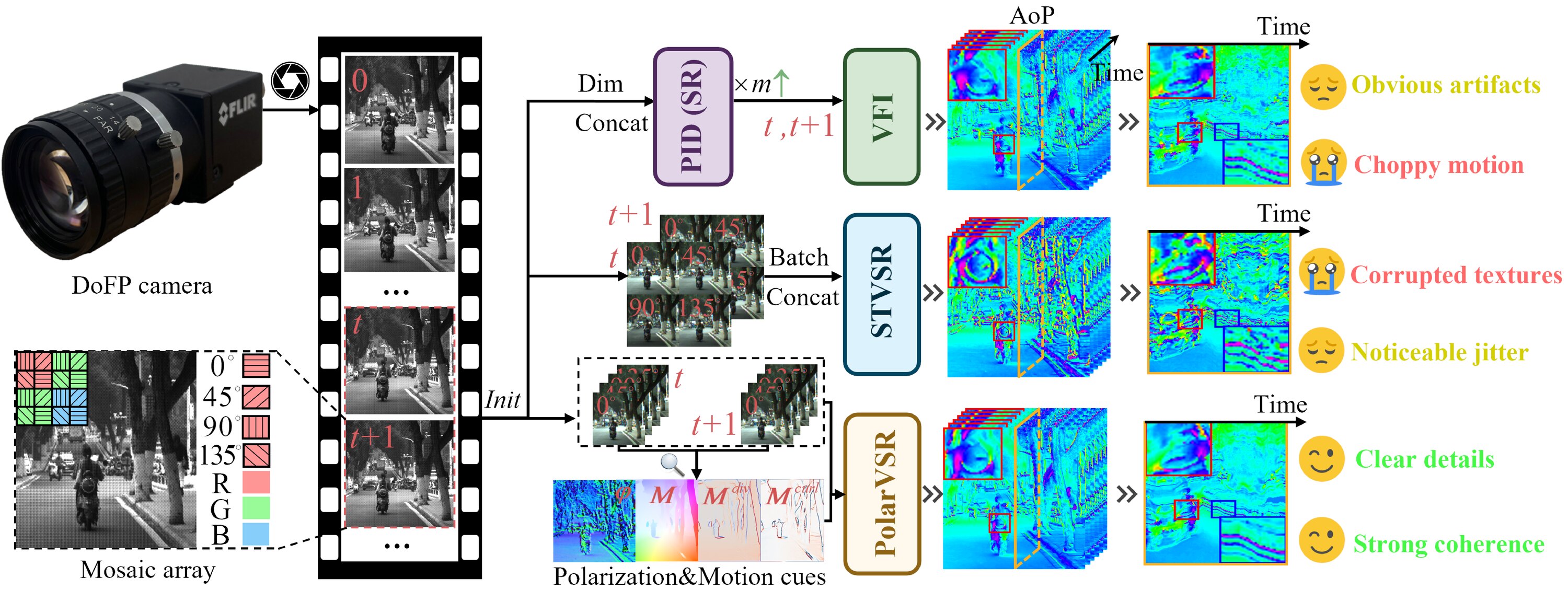}
	\end{center}
	\vspace{-0.8em}
	\caption{The workflow of polarization reconstruction. Video sequences are captured by a DoFP camera, where each mosaic frame samples pixels from four polarization directions following an RGGB order. This work focuses on synthesizing intermediate scenes from two adjacent frames. There are three main paradigms for polarization reconstruction: \textbf{(Top) Cascaded approach}: images first undergo polarization image demosaicking (PID) or super-resolution (SR), followed by video frame interpolation (VFI). This method is prone to error accumulation and lacks space-time interaction. \textbf{(Middle) Space-time video super-resolution (STVSR)}: STVSR methods are applied independently to each polarization direction. This paradigm fails to capture correlations among polarization channels and the temporal evolution of polarization parameters. \textbf{(Bottom) Ours}: the proposed PolarVSR unifies the modeling of all four-directional images. By jointly using motion and polarization cues through tailored modules and losses, the proposed method achieves high-fidelity and temporally consistent reconstruction.}
	\label{fig:1}
\end{figure*}

We first analyze the imaging mechanism of the DoFP color polarization camera (FLIR BFS-U3-51S5PC-C). As shown in Fig. \ref{fig:1},
the sensor integrates a micro-polarizer array with four directions ($0^\circ$, $45^\circ$, $90^\circ$, $135^\circ$) and an RGGB color filter
array. This design captures color information across four polarization directions in one snapshot but causes substantial information loss.
In the mosaic array, only $1/16$ of the original spatial information remains for the red and blue channels, and the green channel preserves
$2/16$ of the data. Thus, high-resolution color polarimetric reconstruction from partial measurements requires effective demosaicking and
super-resolution. Beyond spatial reconstruction, polarization videos require temporal consistency and polarimetric fidelity. The view-dependent nature of polarization \cite{lei2022shape} makes polarimetric parameters vary more strongly than intensity during relative
camera-object motion, complicating video reconstruction. Faced with these challenges, a critical question arises: how can a polarization
reconstruction framework guarantee \textit{polarimetric consistency} throughout the \textit{spatiotemporal} sampling?

Addressing this issue requires a proper treatment of polarimetric dependencies across spatiotemporal dimensions. One intuitive strategy
is a cascaded space-time pipeline (Fig. \ref{fig:1}, top row): frame-wise reconstruction \cite{zhou2025pidsr} is utilized to resolve spatial
polarimetric degradation, and video frame interpolation (VFI) \cite{zhang2025polarization} is then applied to enhance temporal consistency. 
Although direct, this cascaded pipeline isolates spatial and temporal dimensions, preventing mutual interaction. Furthermore, it is computationally
expensive and prone to error accumulation. Another option applies state-of-the-art (SOTA) space-time video
super-resolution (STVSR) methods \cite{chen2022videoinr,chen2023motif,kim2025bf} to each polarization direction independently (Fig. \ref{fig:1},
middle row). These methods usually rely on feature upsampling and warping \cite{jaderberg2015spatial,niklaus2020softmax}. While these frameworks
introduce spatiotemporal modeling, their direct adaptation to polarization reconstruction has two limitations. First, STVSR models are designed for three-channel RGB images, so independent processing cannot
capture inter-direction correlations, and a single network cannot effectively handle degradations across all four polarization directions, leading
to blurred details. Second, STVSR lacks mechanisms and objectives tailored to polarization imaging and therefore neglects the space-time fidelity
of polarization parameters.

To address these difficulties, we develop PolarVSR, the first continuous space-time video reconstruction framework customized for polarization
imaging. We achieve unified spatiotemporal modeling of polarization directions. The framework jointly models the four polarization directions,
promoting the cross-directional interaction of polarimetric properties throughout the spatiotemporal reconstruction. \textbf{Spatially}, PolarVSR facilitates continuous sampling that is 
aligned with polarimetric characteristics. We design a polarization-aware implicit neural representation (PAINR), thereby explicitly encoding both the
polarimetric representations and the physical imaging mechanism into the continuous sampling process. \textbf{Temporally}, PolarVSR utilizes an intensity-guided motion estimation strategy for temporal coherence. 
We introduce a powerful optical flow estimator \cite{wang2024sea} that leverages unpolarized intensity to derive an initial motion estimation
decoupled from polarimetric fluctuations. This initial field is subsequently queried through a space-time INR to yield a high-resolution motion field. Guided by the motion field, the spatial representations are forward-warped \cite{niklaus2020softmax} to align with the target timestamp. To enforce
the \textbf{spatiotemporal} consistency of polarimetric parameters, we deduce the spatiotemporal dynamics of polarization, leveraging the associated
motion and polarimetric cues for feature refinement (MCFR) after feature warping. Moreover, a flow-guided
polarimetric variation loss focuses supervision on polarization-varying regions and helps the network learn spatiotemporal evolution patterns for
consistent polarization reconstruction.

The complex viewpoint dependence of polarization parameters makes synthetic training data insufficient for this task.
Most critically, the community lacks a video-level color polarization benchmark to support space-time polarization video reconstruction (STPVR).
Existing benchmarks generally fall into two categories shown in Table \ref{tab:dataset_comparison}: the first relies on Division-of-Time (DoT) polarimeter systems
\cite{morimatsu2020monochrome,qiu2021linear,wen2019convolutional,li2025demosaicking,zhou2025pidsr,rahman2025polarization}, which offer accurate
color polarization supervision but whose imaging mechanism precludes the acquisition of dynamic video sequences. The second category
uses DoFP cameras, but these are tailored for image-level studies \cite{zhu2024podb,luo2025cpifuse,jeon2024spectral,yao2025polarfree}, limited in
data volume and scene variety, and lack the continuous temporal frames required by STPVR. To fill this gap, we build
the first large-scale DoFP color polarization video benchmark. Captured with a DoFP camera, the benchmark covers diverse materials, motion
patterns, illumination conditions, and indoor and outdoor environments, including challenging real-world scenarios. We also incorporate multiple
denoising strategies to mitigate noise. Together with a training paradigm customized for polarization patterns, this benchmark supports the
development of high-fidelity polarization video reconstruction models.

In summary, the main contributions of this work are as follows:
\begin{itemize}
    \item We formulate space-time polarization video reconstruction and present PolarVSR, the first unified framework for continuous spatiotemporal restoration of polarimetric sequences.
    \item We design a polarization-aware implicit neural representation and an intensity-guided motion estimation strategy to recover spatial details while preserving temporal coherence.
    \item We investigate the spatiotemporal dynamics of polarization parameters to refine and supervise the network for consistent polarization reconstruction.
    \item We build a large-scale real-world color polarization video benchmark to support this research direction.
\end{itemize}

The remainder of this paper is organized as follows. Section II reviews related work. Section III presents the STPVR framework. Section IV describes the benchmark and training paradigm. Section V reports experiments, and Section VI concludes the paper.
\begin{figure*}[!t]
	\begin{center}
		\includegraphics[width=\linewidth]{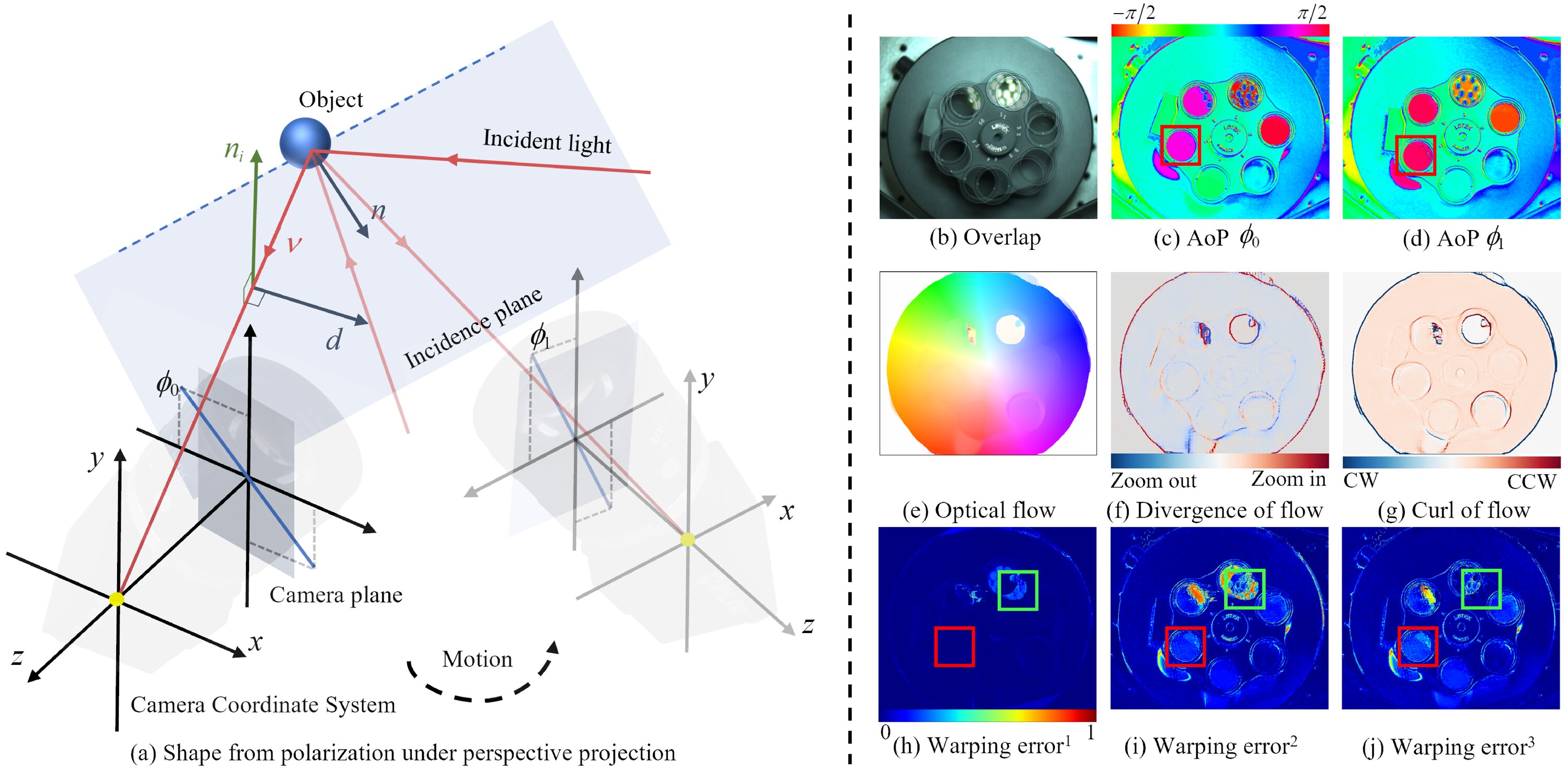}
	\end{center}
		\vspace{-0.8em}
	\caption{The motivation of this work. (a) Shape from polarization under perspective projection. AoP $\phi$ is jointly determined by the viewing direction $\nu $, the polarization direction $d$, and the surface normal $n$. (b) Two overlaid frames showing polarizers rotating on a turntable. (c)-(d) AoP of the previous and next frames, respectively. The AoP of the polarizer changes substantially as the turntable rotates. (e) Optical flow between the two frames estimated from the unpolarized intensity $I$. (f)-(g) Divergence and curl of the optical flow. The signs of divergence and curl denote perspective scaling and rotation, respectively (CW: clockwise, CCW: counter-clockwise). (h)-(i) Errors between the GT and the images $\{I,\phi\}$ warped by optical flow, i.e., $\chi _{I},\chi _{\phi}$ defined in Eq. (\ref{eq:14}). Although optical flow aligns most intensity regions, it cannot fully align AoP because of viewpoint dependence. (j) Warping error between the GT and $\phi$ in well-aligned regions, i.e., $\chi _{AoP}$ defined in Eq. (\ref{eq:15}).}
	\label{fig:2}
\end{figure*}

\begin{table}[!t]
\centering
\caption{Comparison with existing color polarization datasets.}
\label{tab:dataset_comparison}
\footnotesize
\setlength{\tabcolsep}{3pt}
\renewcommand{\arraystretch}{1.05}
\resizebox{\columnwidth}{!}{
\begin{tabular}{lccccc}
\toprule
Dataset & Hardware & Video & Indoor & Outdoor & Images \\
\midrule
Monno \cite{morimatsu2020monochrome} & DoT & -- & \checkmark & \checkmark & 40 \\
Qiu \cite{qiu2021linear} & DoT & -- & \checkmark & -- & 40 \\
Wen \cite{wen2019convolutional} & DoT & -- & \checkmark & \checkmark & 105 \\
DCPM \cite{li2025demosaicking} & DoT & -- & \checkmark & -- & 27 \\
PIDSR \cite{zhou2025pidsr} & DoT & -- & \checkmark & -- & 138 \\
PIDD \cite{rahman2025polarization} & DoT & -- & \checkmark & -- & 160 \\
PODB \cite{zhu2024podb} & DoFP & -- & -- & \checkmark & 24384 \\
CPIFuse \cite{luo2025cpifuse} & DoFP & -- & \checkmark & \checkmark & 100 \\
Jeon \cite{jeon2024spectral} & DoFP & -- & \checkmark & \checkmark & 2022 \\
PolaRGB \cite{yao2025polarfree} & DoFP & -- & \checkmark & \checkmark & 6500 \\
Ours & DoFP & \checkmark & \checkmark & \checkmark & 117550 \\
\bottomrule
\end{tabular}
}
\end{table}

\section{Related Work}
This section reviews representative works on polarization reconstruction and space-time video super-resolution.

\subsection{Polarization Reconstruction}
Polarization reconstruction recovers full-resolution or higher-resolution images across four polarization directions from captured mosaic patterns, corresponding to polarization image demosaicking (PID) and polarization image super-resolution (PISR), respectively. Both tasks estimate polarimetric parameters through spatial resolution enhancement.

Traditional PID methods rely on interpolation \cite{li2019demosaicking, wu2021polarization,morimatsu2020monochrome} or dictionary learning \cite{wen2021sparse, luo2023sparse, luo2024learning}, but often struggle to balance efficiency and accuracy in complex real-world scenarios. With deep learning, Zhang et al. first apply CNNs to PID and reduce over-smoothed outputs using a polarization loss \cite{zhang2018learning}. Sun et al. extend this paradigm to color polarization and contribute a real-world training dataset \cite{sun2021color}. Nguyen et al. reduce learning complexity by decoupling PID into color and polarization demosaicking within a two-stage architecture \cite{nguyen2022two}. Guo et al. design a GAN customized for polarization imaging to synthesize high-frequency polarimetric details \cite{guo2024attention}. More recently, Li et al. introduce diffusion models into PID to improve real-world generalization \cite{li2025demosaicking} and use generative priors from text-to-image models to address data scarcity \cite{li2026pugdiff}.

PISR is typically performed on sub-sampled raw mosaic arrays or sequentially on demosaicked outputs. Hu et al. first introduce a PISR network, formulate a polarization degradation pipeline, and explicitly constrain polarimetric parameters to preserve estimation accuracy \cite{hu2023polarized}. Yu et al. design an intensity-polarization dual-branch network for color polarization super-resolution and use attention to integrate high-frequency textures across domains \cite{yu2023color}. Zhou et al. unify PID and PISR within one architecture and achieve strong performance through joint optimization \cite{zhou2025pidsr}. Mei et al. use event cameras for polarization reconstruction and contribute a large-scale polarization event dataset \cite{mei2023deep}. Hwang et al. derive a polarimetric noise model and establish a robust benchmark for burst polarization super-resolution \cite{hwang2025benchmarking}.

Despite progress in image polarization reconstruction, space-time recovery for polarization videos remains largely unexplored.
Enhancing spatial resolution while preserving temporal consistency of polarimetric parameters is essential for video-level tasks.
Existing methods are also limited to demosaicking or fixed-scale super-resolution. Continuous upsampling typically requires multiple
networks or iterative inference, both lacking flexibility. These limitations motivate a unified space-time polarization reconstruction framework.
\begin{figure*}[!t]
	\begin{center}
		\includegraphics[width=\linewidth]{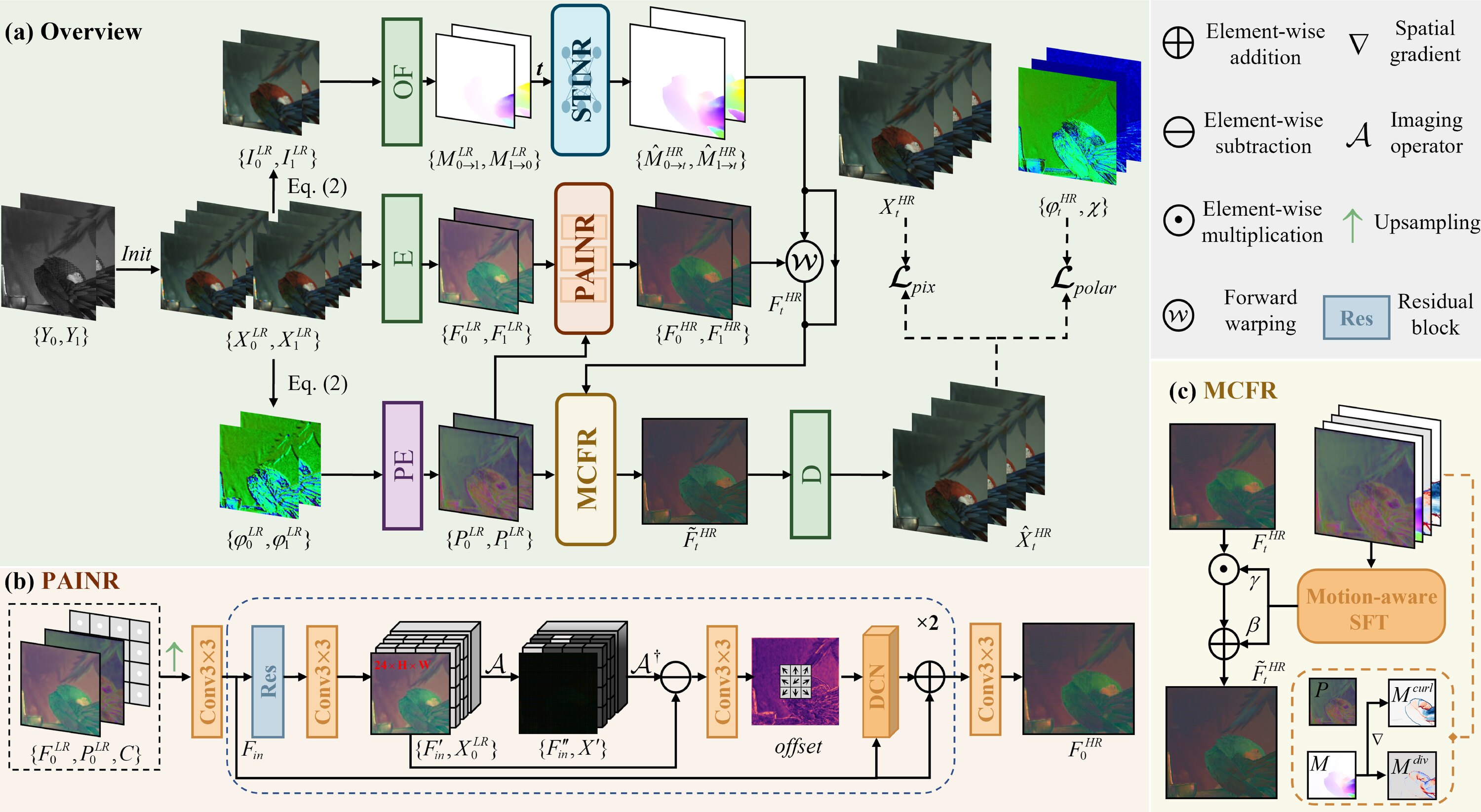}
	\end{center}
		\vspace{-0.8em}
	\caption{Overview of the proposed framework. (a) The workflow. E and D denote the encoder and decoder \cite{chen2023motif},
	respectively. OF is a pre-trained optical flow estimator \cite{wang2024sea}, and PE represents
	the polarization encoder stacked with residual blocks. STINR is adopted from MoTIF \cite{chen2023motif} and consists
	of a series of SIREN layers \cite{sitzmann2020implicit}. The initialized images are first fed into E for feature extraction.
	The features are then upsampled by PAINR and warped \cite{niklaus2020softmax} to the target time with the motion field.
	MCFR is subsequently applied to reduce warping errors. Finally, the refined features are passed through D to reconstruct
	the final image (detailed in Section III-C). (b) Structure of PAINR. It upsamples intensity features $F^{LR}$ by querying the coordinate grid $C$ and polarization features $P^{LR}$. By embedding a forward operator, it defines a discrepancy (Eq. (\ref{eq:8})) to generate deformable convolution offsets and masks for adaptive feature restoration. (c) Structure of MCFR. Designed in an SFT \cite{wang2018recovering} style, the motion-aware SFT uses cascaded residual blocks conditioned on polarization and motion cues. It outputs transformation coefficients to compensate for warping-induced representation errors.}
	\label{fig:3}
\end{figure*}
\subsection{Space-Time Video Super-Resolution}
Instead of treating super-resolution \cite{chan2022basicvsr} and frame interpolation \cite{xu2019quadratic} separately, space-time video super-resolution (STVSR) enhances spatial and temporal resolutions within one framework. Haris et al. establish an early STVSR architecture \cite{haris2020space}. Xiang et al. introduce a bidirectional deformable ConvLSTM to improve one-stage reconstruction \cite{xiang2020zooming}. Xu et al. incorporate temporal modulation for continuous-time super-resolution \cite{xu2021temporal}. More recently, generative priors have entered this domain. He et al. adapt pre-trained text-to-video models for STVSR and show strong robustness against real-world degradations \cite{he2024venhancer}. Concurrently, Wei et al. design a space-time mixture-of-experts with a bidirectional deformable decoder for one-step diffusion STVSR \cite{wei2026osdenhancer}.

However, these frameworks remain restricted to fixed-scale space-time upsampling. VideoINR \cite{chen2022videoinr} pioneers continuous STVSR with implicit neural representations and maintains temporal consistency by warping features with predicted motion fields. Chen et al. replace backward warping in VideoINR with forward warping and incorporate optical flow estimation for more reliable motion prediction \cite{chen2023motif}. Kim et al. improve temporal smoothness with a B-spline mapper and a spatial Fourier mapper for high-frequency textures \cite{kim2025bf}. Becker et al. formulate continuous STVSR as a unified video Fourier field and achieve aliasing-free sampling through a linear point spread function \cite{becker2025continuous}.

Motivated by standard STVSR, we introduce a polarization-customized space-time framework that enhances spatiotemporal resolution while maintaining polarimetric fidelity.

\section{Methodology}
This section formulates polarization imaging and its motivation, presents the network pipeline and modules, and defines the loss functions for optimization.
\subsection{The Mechanism of Polarization Imaging}
Polarization images are captured by integrating micro-polarizers into the camera sensor. The observed image at a given polarization direction is defined as \cite{tzabari2020polarized}:
\begin{equation}
    \begin{split}
{x_\theta } = I\{ 1 + p\cos 2(\theta  - \phi )\} ,
\end{split}
    \label{eq:1}
\end{equation}

where $\theta$ denotes the polarizer direction, $I$ is the unpolarized light intensity, and $p$ and $\phi$ denote the degree and angle of polarization (DoLP and AoP), respectively. These parameters characterize object surface properties. Given captured polarization images, the parameters in Eq. (\ref{eq:1}) are:
\begin{equation}
    \begin{split}
I = \frac{{{S_0}}}{2},p = \frac{{\sqrt {S_1^2 + S_2^2} }}{{{S_0}}},\phi  = \frac{1}{2}\arctan (\frac{{S_2}}{{S_1}}),
\end{split}
    \label{eq:2}
\end{equation}
where $\{ {S_0},{S_1},{S_2}\}$ denotes the Stokes vector \cite{collett2005field}, which can be expressed as:
\begin{equation}
\begin{split}
    S_0 &= \frac{1}{2} ( x_{0^\circ} + x_{45^\circ} + x_{90^\circ} + x_{135^\circ} ), \\
    S_1 &= x_{0^\circ} - x_{90^\circ}, \quad S_2 = x_{45^\circ} - x_{135^\circ}.
\end{split}
\label{eq:3}
\end{equation}

Specifically, $p$ is a function of the surface reflection coefficient and viewing angle \cite{lei2022shape}. As shown in Fig. \ref{fig:2}(a), $\phi$ is governed by the viewing direction $\nu $, polarization direction ${d}$, and surface normal ${n}$:
\begin{equation}
    \begin{split}
\phi  = (d \times \nu ) \times {n_c},
\end{split}
    \label{eq:4}
\end{equation}
where $\times$ denotes the vector product and $n_c$ is the surface normal of the camera plane. The specific formulation of
Eq. (\ref{eq:4}) depends on $d$, which in turn is determined by the reflection type \cite{lei2022shape}.

The preceding analysis shows that, unlike intensity images, polarization parameters $p$ and $\phi$ are highly sensitive to viewpoint changes. As shown in Fig. \ref{fig:2}(b)-(d), intensity images remain relatively consistent during polarizer rotation, whereas $\phi$ exhibits clear numerical shifts. This motion sensitivity requires a tailored polarization video reconstruction pipeline.

To formulate this pipeline, we revisit the imaging operation of the hardware. Referring to Fig. \ref{fig:1}, the DoFP acquisition process is
\begin{equation}
    \begin{split}
Y = {\cal A}{X} +  \eta,
\end{split}
    \label{eq:5}
\end{equation} 
where $Y\in {\mathbb{R}^{1\times H \times W }}$ denotes the mosaic array, ${X} = [ x_{{0^ \circ }},x_{{{45}^ \circ }},x_{{{90}^ \circ }},x_{{{135}^ \circ }}]  \in\mathbb{R} {^{4\times3 \times H \times W}}$ is the full-resolution four-directional color polarization image, and $\eta$ is sensor noise. The operator $\cal A$ samples the color polarization domain by spatially multiplexing $X$ into the single channel $Y$. 


\subsection{Motivation}
As discussed above, polarization parameters show more complex temporal variations under motion than intensity images.
Such dynamics remain unexplored in prior PID and PISR methods, while standard STVSR does not jointly model multiple polarization directions.
These limitations motivate a tailored spatiotemporal reconstruction framework for polarization videos. The method
incorporates polarization cues into the network and uses motion patterns of polarization to refine feature representations.
To support this framework, we further propose a large-scale color polarization video benchmark with diverse scenes and motion patterns for developing and evaluating STPVR models.

\subsection{Overview}
Fig. \ref{fig:3} shows the overall pipeline. Following continuous STVSR frameworks \cite{chen2023motif}, we synthesize
the high-resolution frame $\hat X_t^{HR}$ at arbitrary time $t \in [0,1]$ from adjacent mosaic arrays $\{ {Y_0},{Y_1}\} $.
Taking $t=0$ as an example (the procedure for $t=1$ is identical), the pipeline first processes $Y_0$ with the initialization
in \cite{zhou2025pidsr} to obtain the low-resolution color polarization
image $X_0^{LR}\in\mathbb{R} ^{4\times3 \times \frac{H}{2} \times \frac{W}{2} }$. Unlike standard STVSR,
the method jointly models all four polarization directions. By concatenating $X_0^{LR}$ channel-wise for unified feature
processing (i.e., $4 \times 3 \times \frac{H}{2} \times \frac{W}{2} \to 1 \times 12 \times \frac{H}{2} \times \frac{W}{2}$),
the network learns cross-direction dependencies under degradation, improving polarization parameter estimation.

From $X_0^{LR}$, Eqs. (\ref{eq:2}) and (\ref{eq:3}) compute the unpolarized intensity $I_0^{LR}\in\mathbb{R} ^{1\times3 \times \frac{H}{2} \times \frac{W}{2} }$ and polarization parameters $\varphi _0^{LR}=[\cos 2\phi ,\sin 2\phi ,p]_0^{LR}\in\mathbb{R} ^{1\times9 \times \frac{H}{2} \times \frac{W}{2} }$. An optical flow predictor \cite{wang2024sea} estimates the inter-frame motion field $M_{0 \to 1}^{LR}\in\mathbb{R} ^{1\times2 \times \frac{H}{2} \times \frac{W}{2} }$ from $I_0^{LR}$, while separate encoders extract intensity and polarization representations from $X_0^{LR}$ and $\varphi _0^{LR}$, denoted by $F_0^{LR}\in\mathbb{R} ^{1\times N \times \frac{H}{2} \times \frac{W}{2} }$ and $P_0^{LR} \in\mathbb{R} ^{1\times \frac{N}{2} \times \frac{H}{2} \times \frac{W}{2} }$. PAINR samples $F_0^{LR}$ and $P_0^{LR}$ to produce the high-resolution feature $F_0^{HR}\in\mathbb{R} ^{1\times N \times H \times W }$. Meanwhile, STINR upsamples $M_{0 \to 1}^{LR}$ to the desired time, i.e., $\hat{M}_{0 \to t}^{HR}\in\mathbb{R} ^{1\times 2 \times H \times W }$. Guided by $\{ \hat{M}_{0 \to t}^{HR},\hat{M}_{1 \to t}^{HR}\} $, $\{F_0^{HR},F_1^{HR}\}$ are forward-warped to the target temporal position $F_t^{HR}\in\mathbb{R} ^{1\times N \times H \times W }$. The MCFR block further refines $F_t^{HR}$ to reduce warping-induced polarization artifacts. Finally, a decoder reconstructs $\hat X_t^{HR}\in\mathbb{R} ^{4\times 3 \times H \times W }$ from the refined feature.

\subsection{Upsampling with Implicit Neural Representation}
INR maps discrete encoded features into a continuous feature space and is widely used for continuous STVSR \cite{chen2022videoinr,chen2023motif,kim2025bf}. Following the INR paradigm \cite{chen2021learning}, PAINR and STINR use similar functional forms:
\begin{equation}
    \begin{split}
\begin{array}{l}
F_{{t_r}}^{HR}(z) = {f_{pa}}({v_r},z - {z_r}),\\
M_{{t_r} \to t}^{HR}(z) = {f_{st}}({m_r},z - {z_r},t - {t_r}),
\end{array}
\end{split}
    \label{eq:6}
\end{equation}
where $f_{pa}$ and $f_{st}$ denote the sampling functions of PAINR and STINR, respectively, and $z = (w,h)$ is the target query coordinate. Here, $v_r$ and $m_r$ are latent vectors from the concatenated feature $[F_{tr}^{LR},P_{tr}^{LR}]$ and motion field $M_{0 \to 1}^{LR}$, respectively. These vectors are located at $z_r=(w_r,h_r)$, the nearest coordinate to $z$. Finally, ${t_r} \in \{ 0,1\} $ is the reference temporal index.

Functionally, PAINR spatially upsamples low-resolution intensity representations, whereas STINR outputs the high-resolution motion field at the predicted time. They mainly differ in inputs and sampling functions.

For PAINR, traditional STVSR methods \cite{chen2022videoinr,chen2023motif} rely on MLPs \cite{sitzmann2020implicit} for
pixel-wise feature restoration, which often yields blurred results even under bicubic degradation \cite{wei2026osdenhancer}. Because polarization imaging involves complex cross-channel dependencies and physical modeling, MLP-based functions are less suitable. Inspired by convolutional local image function (CLIF) \cite{chen2024image}, the sampling function $f_{pa}$ of PAINR uses convolutions instead of point-wise MLPs. It treats an image patch as a whole and applies convolutions over the coordinate grid $C$, integrating local context while remaining scale-consistent \cite{chen2024image}. This design improves high-frequency detail restoration during sampling.

The sampling function above focuses on four-directional intensity representations without explicit polarimetric
cues. We therefore embed polarization parameters through a lightweight polarization encoder with five residual blocks.
During continuous sampling, the extracted intensity and polarization embeddings are concatenated,
allowing $f_{pa}$ to aggregate spatially aligned intensity and polarimetric features.

Although the polarization encoder injects polarimetric semantics into PAINR, it does not indicate where DoFP imaging causes severe information loss. Integrating the degradation model benefits inverse problems \cite{wang2022zero,chihaoui2024blind,chung2022diffusion}. Since the imaging operator $\mathcal A$ in Eq. (\ref{eq:5}) is fully known, DoFP imaging is deterministic. We therefore incorporate $\cal A$ into low-resolution feature upsampling. Instead of directly embedding low-quality measurements \cite{saharia2022image} or abstract imaging operators \cite{chen2025invertible},
we use the imaging operator to estimate reconstruction difficulty explicitly, allowing the sampling function to perceive the imaging process and target low-quality regions. Specifically, an intermediate feature $F_{in}$ from $f_{pa}$ is first projected to a 12-channel representation by a convolutional layer. The imaging operation is then applied in this latent space and in the original pixel space:
\begin{equation}
    \begin{split}
\begin{array}{l}
\begin{array}{l}
{{F'}_{in}} = \sigma (Conv({F_{in}})),\\
{{F''}_{in}} = {\cal A}{F'_{in}},X' = {\cal A}{X^{LR}}.
\end{array}
\end{array}
\end{split}
    \label{eq:7}
\end{equation}
Next, a simple pseudo-inverse of $\cal A$ (i.e., bicubic interpolation) is applied to $F''_{in}$ and $X'$. Subtracting the results from $F'_{in}$ and $X^{LR}$ gives the residual:
\begin{equation}
    \begin{split}
\begin{array}{l}
{F_{res}} = {F'_{in}} - {A^\dagger }{F''_{in}},{X_{res}} =  \uparrow (X^{LR} - {A^\dagger }X'),
\end{array}
\end{split}
    \label{eq:8}
\end{equation}
where $\uparrow$ indicates interpolation for spatial alignment. We then use these residuals to guide deformable convolutions (DCN) \cite{zhu2019deformable} for targeted, adaptive feature recovery:
\begin{equation}
    \begin{split}
\begin{array}{l}
o,m = Convs([{F_{res}},{X_{res}}]),\\
{F_{out}} = {F_{in}} + DCN({F_{in}};o,m),
\end{array}
\end{split}
    \label{eq:9}
\end{equation}
where $o$ and $m$ are the learned DCN offsets and modulation masks, respectively, and $Convs$ denotes stacked convolutional layers. Guided by the explicit imaging prior, the adaptive offsets and masks encourage $f_{pa}$ to focus on structurally degraded regions and reduce redundant sampling in flat areas.

For STINR, Eq. (\ref{eq:1}) and Fig. \ref{fig:2}(e)-(g) show that the unpolarized light intensity $I$ is decoupled from polarization parameters and has
limited variation under relative motion. Thus, $\{I^{LR}_0,I^{LR}_1\}$ estimates the low-resolution motion
field $\{M^{LR}_{0 \to 1},M^{LR}_{1 \to 0}\}$ with a pre-trained optical flow network (i.e., SEA-RAFT \cite{wang2024sea}).
Eq. (\ref{eq:6}) then gives the forward high-resolution motion fields $\{\hat{M}^{HR}_{0 \to t},\hat{M}^{HR}_{1 \to t}\}$.
We implement $f_{st}$ as a standard SIREN block \cite{sitzmann2020implicit}. Because motion fields are inherently continuous and smooth,
SIREN is suitable for this task while reducing computational cost.
\subsection{Aligning to the Estimated Time} 
Given high-resolution intensity features $\{F_0^{HR},F_1^{HR}\}$ and motion fields $\{ \hat{M}_{0 \to t}^{HR},\hat{M}_{1 \to t}^{HR}\} $, we synthesize the feature at time $t$ using forward softmax splatting \cite{niklaus2020softmax}:
\begin{equation}
    \begin{split}
\begin{array}{l}
F_t^{HR} = \mathcal{W(}F_0^{HR},F_1^{HR};\hat{M}_{0 \to t}^{HR},\hat{M}_{1 \to t}^{HR}).
\end{array}
\end{split}
    \label{eq:10}
\end{equation}
However, na\"{\i}ve warping introduces additional errors in polarization video reconstruction. As shown in Fig. \ref{fig:2}(h)-(i), the
forward-warped result has a small deviation from the GT for intensity, confirming the effectiveness of warping for
unpolarized light. In contrast, warped AoP shows clear local discrepancies due to its view-dependent characteristics.

To address this issue, MCFR reduces warping artifacts before decoding. MCFR follows the spatial feature transform (SFT) design \cite{wang2018recovering} and generates transformation parameters $\{\gamma,\beta \}$ conditioned on the motion field and polarization features. The motion field implicitly encodes viewpoint cues, and we further formulate divergence and curl as:

\begin{equation}
    \begin{split}
\begin{array}{l}
M^{div} = {\nabla _w}u + {\nabla _h}v,M^{curl} = {\nabla _w}v - {\nabla _h}u,
\end{array}
\end{split} 
    \label{eq:11}
\end{equation}
where $M=(u,v)$ denotes the 2D motion field, and $\nabla _w$ and $\nabla _h$ are first-order differences along the horizontal and vertical axes,
respectively. The superscript ``$HR$'' and temporal subscripts ``$t_{r} \to t$'' are omitted for brevity.
As shown in Fig. \ref{fig:2}(f)-(g), $M^{div}$ and $M^{curl}$ capture local expansion and rotation, thereby
quantifying viewpoint variation. Polarization and motion representations are concatenated as conditions to regress the transformation parameters:
\begin{equation}
    \begin{split}
\begin{array}{l}
\begin{array}{l}
\gamma ,\beta  = Convs(P,M'),\tilde F_t^{HR} = (1+\gamma)  \odot F_t^{HR} + \beta ,
\end{array}
\end{array}
\end{split}
    \label{eq:12}
\end{equation}
where $P = [P_0^{HR},P_1^{HR}]$ denotes upsampled polarization features, $M' = [M,{M^{div}},{M^{curl}}]$ denotes the motion representation,
and $\odot$ is the Hadamard product. $\tilde F_t^{HR}$ is then concatenated with polarization and motion features and fed into the decoder for final reconstruction.

\begin{figure}[!t]
	\begin{center}
		\includegraphics[width=\linewidth]{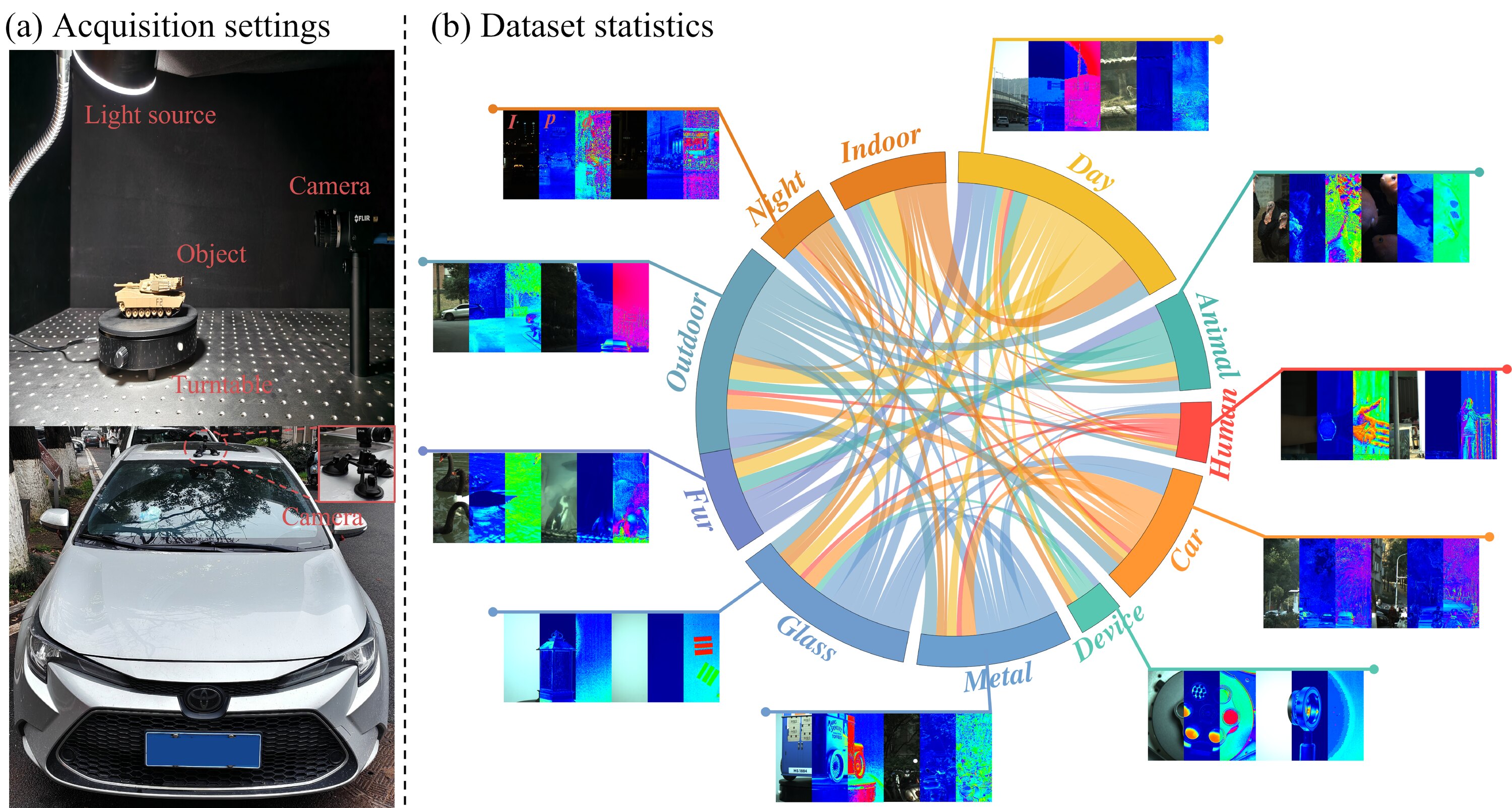}
	\end{center}
	\vspace{-0.8em}
	\caption{Overview of the proposed benchmark. (a) Acquisition settings. Indoor moving objects are captured using a workbench
	and a turntable, whereas outdoor sequences are captured via tripods or vehicle-mounted cameras. (b) Dataset statistics.
	${I,p,\phi}$ are jointly visualized. The chord diagram illustrates the rich scene variety within our benchmark.}
	\label{fig:4}
\end{figure}

\begin{figure}[!t]
	\begin{center}
		\includegraphics[width=\linewidth]{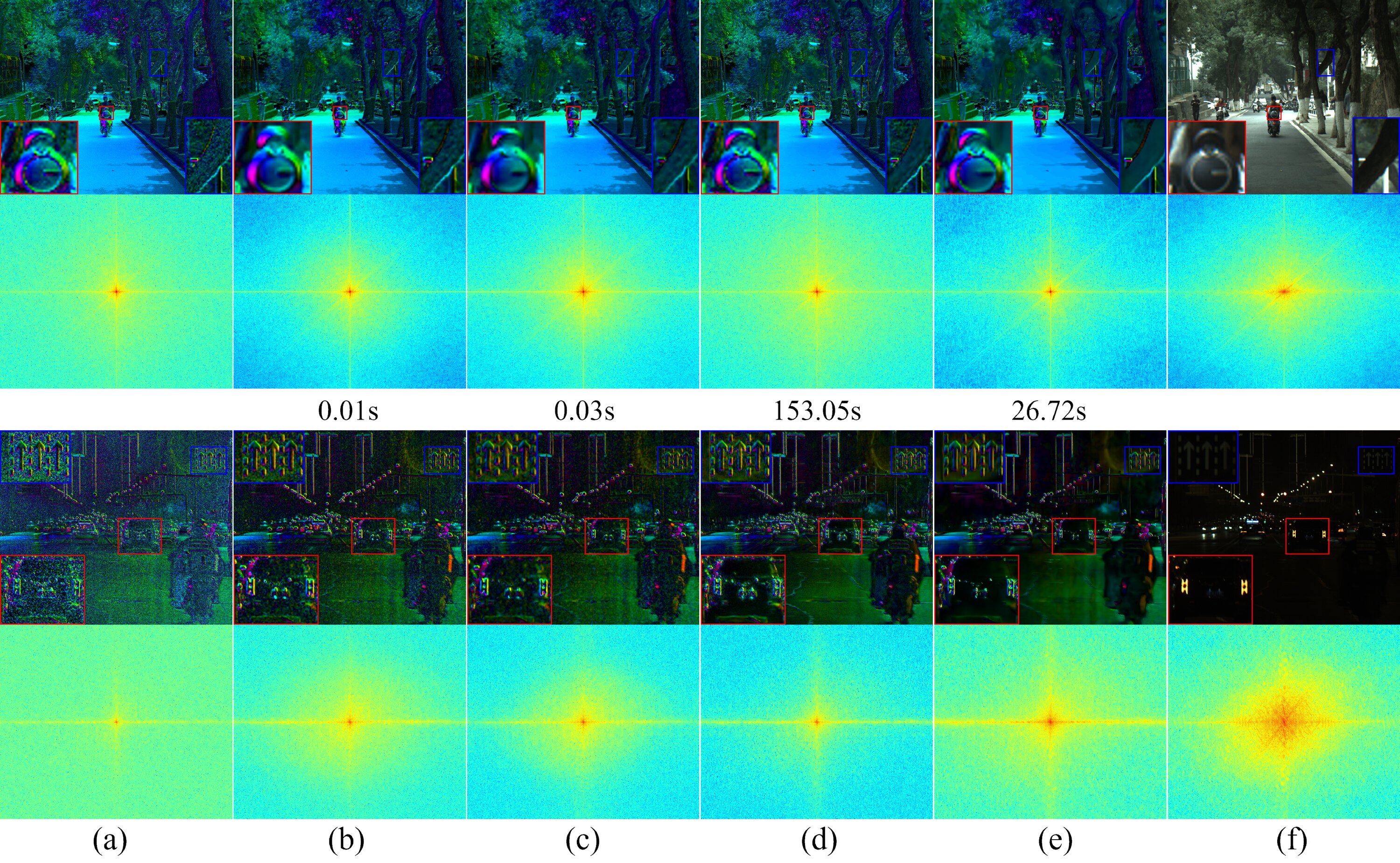}
	\end{center}
	\vspace{-0.8em}
	\caption{Visualization of different denoising strategies. The first and third rows display $p$ and $\phi$, while the second and fourth rows show the corresponding FFT spectra. The reported time denotes the processing time for a single $400 \times 400$ polarization image. (a) Original image. (b) Gaussian filtering. (c) Guided filtering. (d) PCA denoising. (e) BM3D denoising. (f) $I$ of the scene.
	}
	\label{fig:5}
\end{figure}


\subsection{Loss Function}
The loss function contains three terms. The first two constrain the four-directional intensity images and predicted motion field:
\begin{equation}
    \begin{split}
\begin{array}{l}
{{\cal L}_{int}} = \sqrt {{{\left\| {\hat X_t^{HR} - X_t^{HR}} \right\|}^2} + {\varepsilon ^2}} ,\\
{{\cal L}_{flow}} = \sum\limits_{{t_r} = 0}^1 {\sqrt {{{\left\| {\hat M_{{t_r} \to t}^{HR} - M_{{t_r} \to t}^{HR}} \right\|}^2} + {\varepsilon ^2},} } 
\end{array}
\end{split}
    \label{eq:13}
\end{equation}
where $\hat X_t^{HR}$ and $\hat M_{{t_r} \to t}^{HR}$ denote predictions, $X_t^{HR}$ and $ M_{{t_r} \to t}^{HR}$ denote the GT, and $\varepsilon$ is empirically set to $10^{-5}$.

Intensity supervision alone yields suboptimal results and fails to capture spatiotemporal dynamics in polarization parameters.
Therefore, guided by the motion cues, we identify regions exhibiting spatiotemporal polarimetric variations. This directs the network to attend to
the polarimetric evolution within these dynamic areas. Formally, we first define the mask as:
\begin{equation}
    \begin{split}
\begin{array}{l}
\tilde V_{t}^{HR} = \mathcal{W}(V_{t-1}^{HR},M_{t-1 \to t }^{HR}),{\chi _V} = \mathcal{N}(\mathcal{D}_V(V_{t }^{HR},\tilde V_{t}^{HR})),
\end{array}
\end{split}
    \label{eq:14}
\end{equation}
where $V \in \{ I,p,\phi\}$, and $M^{HR}_{t-1 \to t}$ is the optical flow derived from $\{I^{HR}_{t-1},I^{HR}_{t}\}$ to warp $V_{t-1}^{HR}$ to $\tilde V_{t}^{HR}$.
The distance between $\tilde V_{t}^{HR}$ and the corresponding GT is computed by $\mathcal D_V$. In implementation, $\mathcal D_I$
and $\mathcal D_p$ are the Charbonnier losses in Eq. (\ref{eq:13}), and $\mathcal D_\phi$ is a cosine loss for AoP periodicity.
The distance is normalized by $\mathcal N$ to generate the mask $\chi _V$ shown in Fig. \ref{fig:2}(h)-(j),
indicating temporal transitions of polarization properties at corresponding spatial locations. We further define:
\begin{equation}
    \begin{split}
\begin{array}{l}
{\chi _{DoLP}} = {e^{ - \tau {\chi _I}}}{\chi _p},{\chi _{AoP}} = {e^{ - \tau {\chi _I}}}{\chi _\phi },
\end{array}
\end{split}
    \label{eq:15}
\end{equation}
where $e$ denotes the exponential function, and $\tau$ is the decay rate set to $10$. Eq. (\ref{eq:15}) activates
the polarization mask only in well-aligned intensity regions, as shown in Fig. \ref{fig:2}(j). The flow-guided polarization variation loss is defined as:
\begin{equation}
    \begin{split}
\begin{array}{l}
{{\cal L}_{{\mathop{\rm var}} }} = {\chi _{DoLP}}\mathcal{D}_p(\hat p_t^{HR},p_t^{HR}) + {\chi _{AoP}}\mathcal{D_\phi}(\hat \phi _t^{HR},\phi _t^{HR}).
\end{array}
\end{split}
    \label{eq:16}
\end{equation}

 Unlike Section III-E, the mask in Eq. (\ref{eq:14}) is computed entirely from GT frames and therefore represents true polarization progression. Incorporating this mask into the loss encourages the network to focus on regions with polarization shifts and learn polarization evolution patterns.
 
 For the remaining smooth regions, we introduce a loss inspired by \cite{lai2018learning} to improve temporal coherence:
 \begin{equation}
    \begin{split}
\begin{array}{l}
{{\chi '}_{DoLP}} = {e^{ - \tau {\chi _p}}},{{\chi '}_{AoP}} = {e^{ - \tau {\chi _\phi }}},\\
{{\cal L}_{sm}} = {{\chi '}_{DoLP}}D(\hat p_t^{HR},\tilde p_t^{HR}) + {{\chi '}_{AoP}}D(\hat \phi _t^{HR},\tilde \phi _t^{HR}),
\end{array}
\end{split}
    \label{eq:17}
\end{equation}
where $\tilde p _t^{HR}$ and $\tilde \phi _t^{HR}$ are obtained by Eq. (\ref{eq:14}), and the total loss is:
 \begin{equation}
    \begin{split}
\begin{array}{l}
{{\cal L}_{pix}} = {{\cal L}_{{\rm{ }}int}} + {\lambda _1}{{\cal L}_{flow}},{{\cal L}_{polar}} = {{\cal L}_{{\rm{var}}}} + {{\cal L}_{sm}},\\
{{\cal L}_{total}} = {{\cal L}_{pix}} + {\lambda _2}{{\cal L}_{polar}},
\end{array}
\end{split}
    \label{eq:18}
\end{equation}
where $\lambda _1$ and $\lambda _2$ are balancing hyperparameters.

\section{Benchmark and Training Pipeline}
\subsection{Benchmark Overview}
We introduce PV, the first large-scale color polarization video dataset collected with a FLIR BFS-U3-51S5PC-C camera.
As shown in Fig. \ref{fig:4}, PV covers diverse indoor and outdoor scenes. Indoors, the camera is mounted on a rigid stand,
and objects are placed on a motorized turntable. The turntable uses three speed levels, and light sources are adjusted for different
illumination conditions. The objects span materials including polarizers, plastics, and frosted surfaces. Outdoors, some
scenes are recorded using a tripod (e.g., in a zoo), and we also collect in-vehicle driving sequences under daytime and nighttime road conditions.
The camera operates at 75 FPS. In total, the dataset contains 65 scenes and 117550 frames, with sequence lengths from 200 to 2000 frames.

\subsection{Denoising Strategies}
For polarization video reconstruction, imaging quality is lower than that of a DoT polarimeter system,
and frame averaging cannot suppress noise in the DoFP system. To smooth polarization parameters, we consider two efficient denoising strategies,
Gaussian filtering \cite{koenderink1984structure} and guided filtering \cite{he2012guided},
and two resource-intensive strategies, PCA denoising \cite{zhang2017pca} and BM3D \cite{rahman2025polarization}.

Both Gaussian and guided filtering are performed on the $S_1$ and $S_2$ components. PCA denoising and BM3D also consider correlations between polarization channels, as described in their original papers \cite{zhang2017pca,rahman2025polarization}. For guided filtering, we have:
 \begin{equation}
    \begin{split}
{{\tilde S}_1},{{\tilde S}_2} = GF({S_1},{S_2};I),
\end{split}
    \label{eq:19}
\end{equation}
$p$ and $\phi$ are then computed from ${{\tilde S}_1},{{\tilde S}_2}$ using Eq. (\ref{eq:2}). Eq. (\ref{eq:19}) follows the
observation that the intensity image $I$ contains rich textures and has low noise under the capture conditions.
Thus, $I$ is a suitable guidance image for filtering polarization parameters.

Fig. \ref{fig:5} compares denoising strategies. In the daytime scene, all strategies suppress noise effectively. With high-quality guidance images, guided filtering preserves textures better than Gaussian filtering and retains more high-frequency spectral components. BM3D removes noise most strongly but over-smooths details. PCA denoising gives the best overall result by preserving polarization details while filtering noise, but its high computational cost is impractical for large-scale video datasets. In nighttime scenes, Gaussian filtering has the weakest denoising capability, and guided filtering also fails because the guidance image is low quality. In contrast, BM3D and PCA denoising still perform well. The spectra show clear directionality in the remaining high-frequency components, benefiting from their strong noise modeling capabilities.

Based on these observations, we use efficient guided filtering for high-luminance scenes, which constitute most of the benchmark. For challenging nighttime scenes, we apply PCA denoising and BM3D to ensure the quality of polarization parameters. All four
strategies will be implemented in our project. These denoising strategies are used only for training
supervision. The model input remains unprocessed raw mosaic frames, ensuring consistency with inference.
\subsection{Degradation Pipeline}
The training pipeline largely follows the continuous STVSR paradigm \cite{chen2022videoinr,chen2023motif,kim2025bf}, with the main difference being the degradation. In PID and PISR, the GT can be obtained using a DoT polarimeter system.
However, collecting DoT-based GT is infeasible for DoFP video sequences.
We therefore formulate a practical workaround for supervised learning: the mosaic array is reorganized into $4 \times $ downsampled
color polarization images ${}^{4 \downarrow }{X^{HR}}$ as proxy GT ($4H \times 4W \to 4 \times 3 \times H \times W$, where G1 and G2
channels are averaged). During training, ${}^{4 \downarrow }{X^{HR}}$ is first downsampled
to ${}^{4m \downarrow }{X^{HR}\in\mathbb{R} {^{4\times3 \times \frac{H}{m} \times \frac{W}{m}}}}$ $(m\ge1)$ by bicubic interpolation.
Then, the imaging operator $\mathcal A$ processes ${}^{4m \downarrow }{X^{HR}}$ to generate the network
input $X^{LR}\in\mathbb{R} {^{4\times3 \times \frac{H}{2m} \times \frac{W}{2m}}}$. Thus, ${}^{4 \downarrow }{X^{HR}}$
and $X^{LR}$ form a valid training pair. More details are provided in the supplementary material.

\begin{figure*}[!t]
	\begin{center}
		\includegraphics[width=\linewidth]{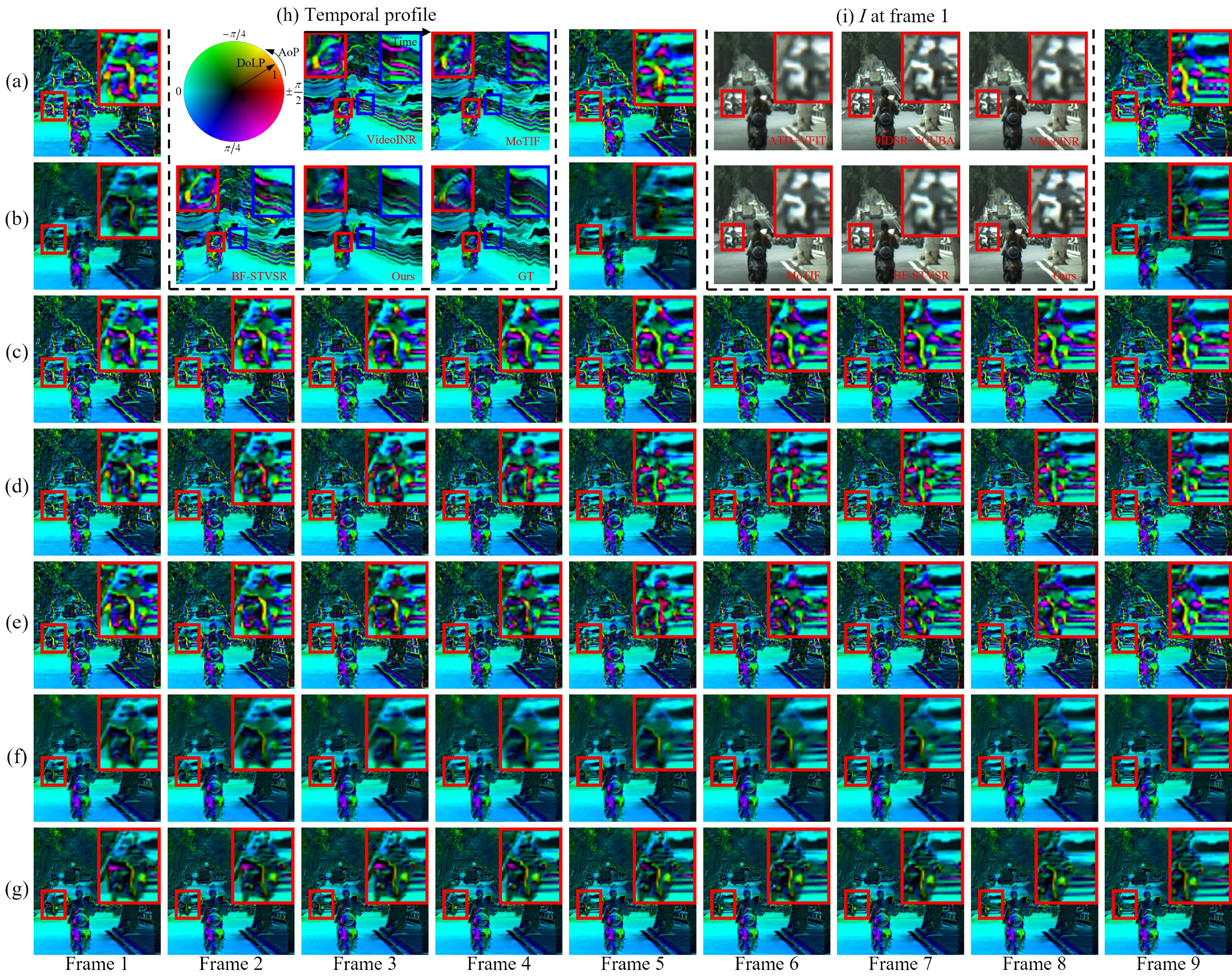}
	\end{center}
	\vspace{-0.8em}
	\caption{Visual comparisons of demosaicking ($2 \times$) and $8\times$ interpolation results on synthetic data. The horizontal and vertical axes represent temporal frames and methods, respectively. DoLP and AoP are jointly visualized using the HSV color space shown in (h). (a)-(b) Results of ATD \cite{zhang2026atd}+VFIT \cite{lu2022video} and PIDSR \cite{zhou2025pidsr}+SCUBA \cite{zhang2025polarization}, respectively. Since these VFI algorithms only support $2 \times$ upsampling, most intermediate timesteps are left blank. (c)-(e) Results of VideoINR \cite{chen2022videoinr}, MoTIF \cite{chen2023motif}, and BF-STVSR \cite{kim2025bf}, which are continuous STVSR methods. (f) Ours. (g) GT. (h) Temporal profiles. From left to right and top to bottom: VideoINR, MoTIF, BF-STVSR, ours, and GT. (i) The unpolarized light $I$ at frame 1. From left to right and top to bottom: ATD+VFIT, VideoINR, MoTIF, BF-STVSR, and ours.}
	\label{fig:7}
\end{figure*}
\section{Experiments}
\subsection{Implementation and Training Details}
We evaluate the network on the proposed benchmark, which is divided into 122 clips, with 70 for
training and 52 for testing. Following previous works \cite{chen2022videoinr,chen2023motif,kim2025bf}, training
uses two stages of 150k steps each. In the first stage, the imaging operator is
applied without downsampling (i.e., $m=1$), so the network learns only demosaicking ($2\times$). To optimize
joint demosaicking and super-resolution, the second stage first downsamples images by
$m \in \{ 1,2,4\} $ and then applies imaging degradation ($2\times$, $4\times$, $8\times$).
$\lambda_1$ and $\lambda_2$ in Eq. (\ref{eq:18}) are set to 0.1 and 0.2, respectively. The optimizer and scheduler settings follow \cite{chen2023motif}. We use a batch size of 4 and randomly crop images to $128\times128$ for augmentation. All experiments are conducted on NVIDIA RTX 4090 GPUs.
\begin{table}[t] 
\centering
\caption{Quantitative comparison of different methods under the demosaicking ($2\times$) and $2\times$ frame interpolation setting. 
The best and second-best results are marked in \textbf{bold} and \underline{underline}, respectively.}
\label{tab:1}
\renewcommand{\arraystretch}{1.05}
\setlength{\tabcolsep}{3pt} 
\resizebox{\columnwidth}{!}{ 
\begin{tabular}{llccccc}
\toprule
\makecell{VFI\\Method} & \makecell{SR\\Method} & PSNR$_{I}\uparrow$ & PSNR$_{p}\uparrow$ & SSIM$_{I}\uparrow$ & SSIM$_{p}\uparrow$ & MAE$\downarrow$ \\
\midrule

\multirow{3}{*}{SuperSloMo \cite{jiang2018super}}
& ATD \cite{zhang2026atd}      & 27.322 & 23.073 & 0.847 & 0.657 & 14.626 \\
& PIDSR \cite{zhou2025pidsr}    & 33.977 & 32.021 & 0.939 & 0.818 & 11.843 \\
& PUGDiff \cite{li2026pugdiff}  & 28.039 & 30.299 & 0.862 & 0.781 & 8.784 \\
\midrule

\multirow{3}{*}{VFIT \cite{lu2022video}}
& ATD      & 28.983 & 22.390 & 0.875 & 0.658 & 14.693 \\
& PIDSR    & \underline{34.052} & \underline{32.343} & \underline{0.940} & \underline{0.829} & 11.443 \\
& PUGDiff  & 30.225 & 31.043 & 0.897 & 0.800 & 8.907 \\
\midrule

\multirow{3}{*}{SCUBA \cite{zhang2025polarization}}
& ATD      & 28.941 & 22.585 & 0.875 & 0.657 & 15.106 \\
& PIDSR    & 30.471 & 31.289 & 0.894 & 0.796 & 12.769 \\
& PUGDiff  & 30.212 & 30.876 & 0.896 & 0.791 & 9.995 \\
\midrule

\multicolumn{2}{c}{VideoINR \cite{chen2022videoinr}} & 29.388 & 22.218 & 0.886 & 0.657 & 11.255 \\

\multicolumn{2}{c}{VideoINR-12ch} & 32.423 & 29.732 & 0.924 & 0.771 & 8.583 \\

\multicolumn{2}{c}{MoTIF \cite{chen2023motif}}    & 29.687 & 22.519 & 0.892 & 0.660 & 11.415 \\

\multicolumn{2}{c}{MoTIF-12ch} & 32.292 & 29.859 & 0.933 & 0.788 & 8.558 \\

\multicolumn{2}{c}{BF-STVSR \cite{kim2025bf}} & 29.472 & 22.214 & 0.890 & 0.656 & 11.481 \\

\multicolumn{2}{c}{BF-STVSR-12ch} & 32.654 & 29.763 & 0.928 & 0.776 & \underline{8.437} \\
\midrule
\multicolumn{2}{c}{Ours}     & \textbf{34.631} & \textbf{33.310} & \textbf{0.944} & \textbf{0.854} & \textbf{5.922} \\
\bottomrule
\end{tabular}
}
\end{table}
\begin{figure*}[!t]
	\begin{center}
		\includegraphics[width=\linewidth]{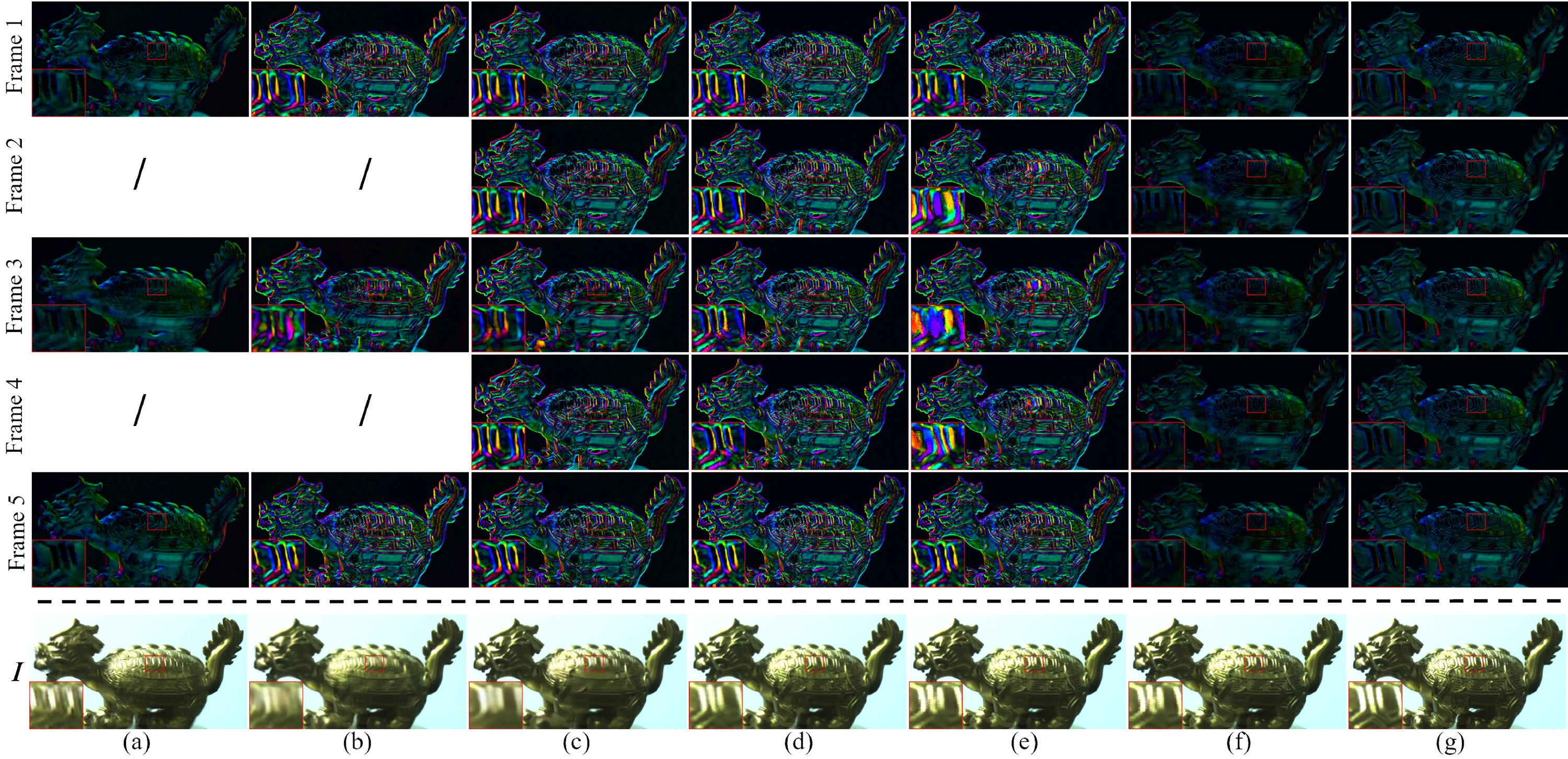}
	\end{center}
	\vspace{-0.8em}
\caption{Visual comparisons of $4\times$ super-resolution and $4\times$ interpolation results on real-world data. The horizontal and vertical axes represent methods and temporal frames, respectively. DoLP and AoP are jointly visualized. The last row displays the unpolarized light $I$ at frame 3. (a)-(b) Results of PIDSR \cite{zhou2025pidsr}+SCUBA \cite{zhang2025polarization} and ZoomingSlowMo \cite{xiang2020zooming}, respectively. Since these interpolation patterns only support $2 \times$ upsampling, some intermediate timesteps are left blank. (c)-(f) Results of TMNet \cite{xu2021temporal}, MoTIF \cite{chen2023motif}, BF-STVSR \cite{kim2025bf}, and BF-STVSR-12ch, respectively. (g) Ours.}
	\label{fig:8}
\end{figure*}
\begin{figure*}[!t]
	\begin{center}
		\includegraphics[width=\linewidth]{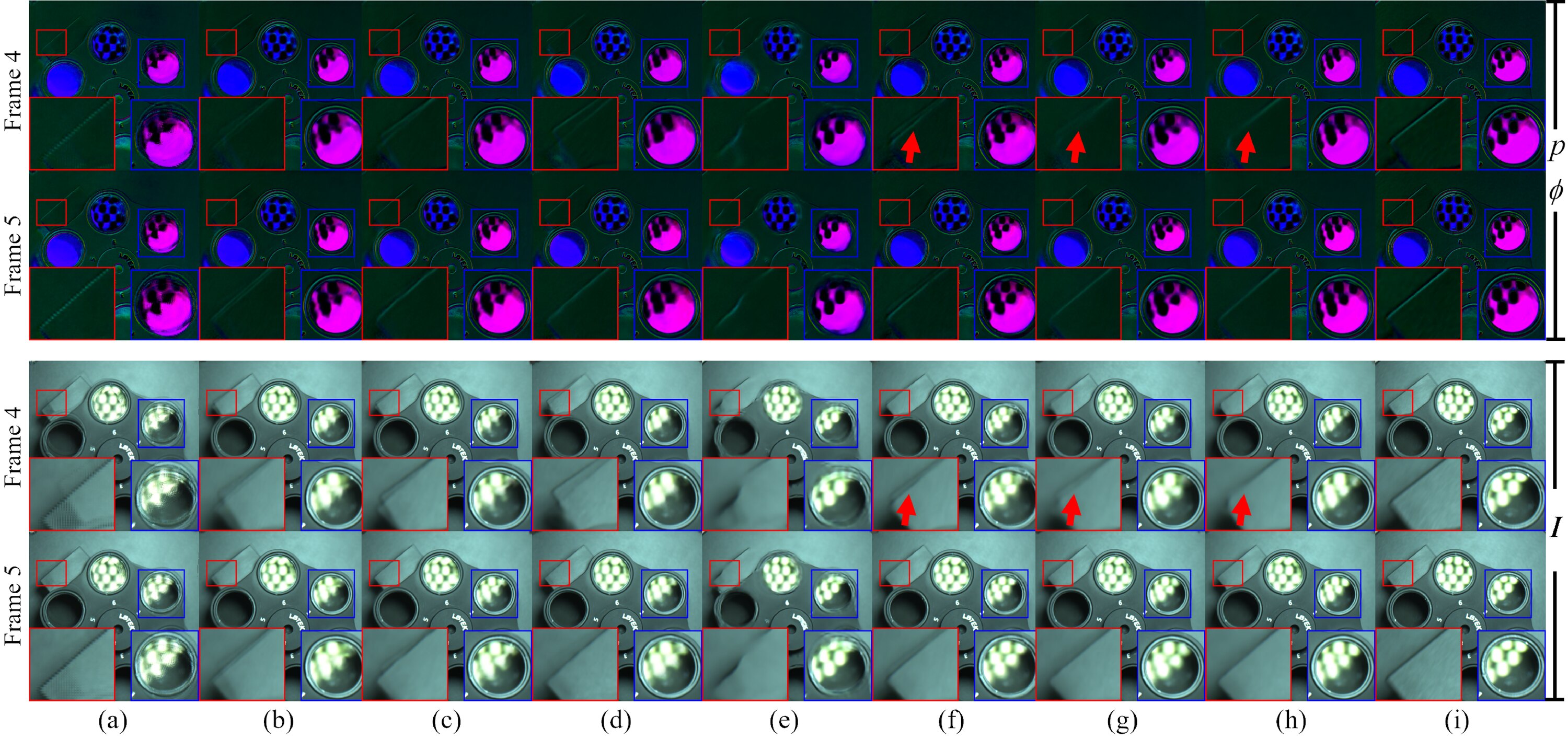}
	\end{center}
	\vspace{-0.8em}
	\caption{Visualizations of the ablation study on synthetic data for demosaicking ($2 \times$) and $8 \times$ interpolation. The horizontal and vertical axes represent different configurations and temporal frames, respectively. (a) Replacing CLIF with SIREN in $f_{pa}$. (b) W/o polarization injection. (c) W/o $\mathcal{A}$ in $f_{pa}$. (d) W/o MCFR. (e) W/o the optical flow estimator. (f) W/o ${{\cal L}_{polar}}$. (g) W/o $\chi$ in ${{\cal L}_{polar}}$. (h) Full configuration. (i) GT.}
	\label{fig:10}
\end{figure*}
\subsection{Baselines and Evaluation Metrics}
We compare the proposed method with two paradigms. The first uses a two-step approach,
where spatial upsampling (demosaicking and SR) is followed by VFI. For spatial upsampling, we use
PIDSR \cite{zhou2025pidsr}, PUGDiff \cite{li2026pugdiff}, and ATD \cite{zhang2026atd}; for VFI,
we use SCUBA \cite{zhang2025polarization}, VFIT \cite{lu2022video}, and SuperSloMo \cite{jiang2018super}.
PIDSR, PUGDiff, and SCUBA are tailored for polarization images, whereas the other methods are trained
for conventional RGB images. The second consists of standard STVSR methods, including
ZoomingSlowMo \cite{xiang2020zooming}, TMNet \cite{xu2021temporal}, VideoINR \cite{chen2022videoinr},
MoTIF \cite{chen2023motif}, and BF-STVSR \cite{kim2025bf}. The latter three support continuous space-time upsampling.
All these methods target conventional RGB images. To adapt RGB baselines to degradation patterns
of polarization imaging, we fine-tune their weights on the proposed benchmark. Specifically, the four-directional
intensity images are concatenated along the batch dimension to fit the RGB channel. Additionally, we retrain the baselines
by adjusting their input and output channels to 12, enabling them to model the connections among polarization directions.
These retrained models are denoted as ``VideoINR-12ch'', ``MoTIF-12ch'' and ``BF-STVSR-12ch''.

For metrics, we use \text{PSNR} and \text{SSIM} to assess unpolarized light intensity $I$ and DoLP $p$, reported as \text{PSNR$_I$}, \text{SSIM$_I$}, \text{PSNR$_p$}, and \text{SSIM$_p$}. For AoP $\phi$, we use mean angular error (\text{MAE}) \cite{li2026pugdiff}.

As discussed in Section IV, acquiring GT video pairs for quantitative evaluation is infeasible. We therefore synthesize test pairs using the degradation pipeline for metric calculation and provide qualitative results on real-world mosaic arrays. These two evaluation scenarios are denoted as ``synthetic'' and ``real-world'', respectively.
\subsection{Comparison with Baselines}
\textbf{Quantitative evaluation.} Tables \ref{tab:1} and \ref{tab:2} report results of different methods on the
proposed benchmark for joint VFI ($2\times,4\times,8\times$) with demosaicking ($2\times$) and SR ($4\times,8\times$).
For demosaicking (Table \ref{tab:1}), PIDSR significantly outperforms three-channel RGB models in terms of $I$ and $p$.
This suggests that three-channel RGB models fail to model connections among polarization directions in polarization
video reconstruction. Modeling each direction independently during training leads to averaged results and suboptimal metrics.
The 12-channel baselines surpass their 3-channel counterparts and reach performance comparable to the PID model,
further demonstrating the necessity of joint polarization modeling.

For higher-scale space-time SR (Table \ref{tab:2}), PIDSR cannot match STVSR models because of
architectural limitations, while ZoomingSlowMo and TMNet do not support arbitrary scale factors. This comparison
shows the effectiveness of continuous STVSR. By using polarization and motion cues within this paradigm, the proposed method introduces customized modules and losses for accurate polarization parameter estimation. It therefore outperforms continuous STVSR models and 12-channel baselines
by a large margin across all metrics and shows clear advantages over PID and PISR methods.


\begin{table*}[t]
\centering
\caption{Quantitative super-resolution comparison under different spatial-temporal settings. 
Each entry reports $\mathrm{PSNR}_{I} / \mathrm{PSNR}_{p} / \mathrm{MAE}$. 
The best and second-best results for each metric under each setting are marked in \textbf{bold} and \underline{underline}, respectively.}
\label{tab:2}
\renewcommand{\arraystretch}{1.03}
\setlength{\aboverulesep}{0.35ex}
\setlength{\belowrulesep}{0.35ex}
\setlength{\tabcolsep}{5.5pt}
\resizebox{\textwidth}{!}{
\begin{tabular}{lcccc}
\toprule
Method & Spatial Scale & Time $\times 2$ & Time $\times 4$ & Time $\times 8$ \\
\midrule

\multirow{2}{*}{SuperSloMo \cite{jiang2018super}+PIDSR \cite{zhou2025pidsr}}
& $\times 4$ & 21.469 / 24.755 / 16.034 & -- & -- \\
& $\times 8$ & 20.044 / 21.681 / 18.005 & -- & -- \\
\midrule

\multirow{2}{*}{VFIT \cite{lu2022video}+PIDSR}
& $\times 4$ & 21.642 / 24.117 / 15.639 & -- & -- \\
& $\times 8$ & 20.082 / 21.268 / 17.821 & -- & -- \\
\midrule

\multirow{2}{*}{SCUBA \cite{zhang2025polarization}+PIDSR}
& $\times 4$ & 21.695 / 24.240 / 15.615 & -- & -- \\
& $\times 8$ & 20.101 / 21.345 / 17.784 & -- & -- \\
\midrule

\multirow{2}{*}{ZoomingSlowMo \cite{xiang2020zooming}}
& $\times 4$ & 25.264 / 19.284 / 17.663 & -- & -- \\
& $\times 8$ & -- & -- & -- \\
\midrule

\multirow{2}{*}{TMNet \cite{xu2021temporal}}
& $\times 4$ & 25.279 / 19.456 / 17.058 & 24.916 / 19.463 / 17.190 & 24.022 / 19.357 / 17.653 \\
& $\times 8$ & -- & -- & -- \\
\midrule

\multirow{2}{*}{VideoINR \cite{chen2022videoinr}}
& $\times 4$ & 25.557 / 19.195 / 14.659 & 25.256 / 19.330 / 14.636 & 24.407 / 19.488 / 14.955 \\
& $\times 8$ & 22.162 / 17.053 / 18.347 & 22.096 / 17.211 / 18.187 & 21.820 / 17.324 / 18.304 \\
\midrule

\multirow{2}{*}{VideoINR-12ch}
& $\times 4$ & 27.161 / 27.986 / 10.297 & 26.621 / 27.880 / 10.424 & 25.367 / 27.523 / 10.768 \\
& $\times 8$ & 23.353 / 26.184 / 12.065 & 23.223 / 26.190 / \underline{12.015} & 22.716 / 26.090 / 12.133 \\
\midrule

\multirow{2}{*}{MoTIF \cite{chen2023motif}}
& $\times 4$ & 25.651 / 19.438 / 14.472 & 25.346 / 19.537 / 14.444 & 24.491 / 19.546 / 14.741 \\
& $\times 8$ & 22.120 / 16.955 / 18.203 & 22.039 / 17.135 / 18.049 & 21.737 / 17.211 / 18.178 \\
\midrule

\multirow{2}{*}{MoTIF-12ch}
& $\times 4$ & \underline{28.165} / \underline{28.429} / 10.257 & \underline{27.401} / \underline{28.332} / 10.343 & \underline{25.731} / \underline{28.003} / 10.653 \\
& $\times 8$ & \underline{24.059} / \underline{26.553} / 12.410 & 23.565 / 26.184 / 12.015 & 22.905 / \underline{26.416} / 12.533 \\
\midrule

\multirow{2}{*}{BF-STVSR \cite{kim2025bf}}
& $\times 4$ & 25.528 / 19.382 / 14.396 & 25.263 / 19.480 / 14.373 & 24.488 / 19.511 / 14.659 \\
& $\times 8$ & 22.105 / 16.952 / 18.118 & 22.054 / 17.099 / 17.973 & 21.799 / 17.160 / 18.089 \\
\midrule

\multirow{2}{*}{BF-STVSR-12ch}
& $\times 4$ & 27.698 / 28.068 / \underline{10.236} & 27.083 / 27.970 / \underline{10.319} & 25.703 / 27.661 / \underline{10.562} \\
& $\times 8$ & 23.728 / 26.193 / \underline{12.060} & \underline{23.688} / \underline{26.551} / 12.395 & \underline{22.979} / 26.091 / \underline{12.100} \\
\midrule

\multirow{2}{*}{Ours}
& $\times 4$ & \textbf{30.104} / \textbf{30.997} / \textbf{7.285} & \textbf{29.084} / \textbf{30.753} / \textbf{7.415} & \textbf{27.054} / \textbf{30.098} / \textbf{7.800} \\
& $\times 8$ & \textbf{25.086} / \textbf{28.497} / \textbf{9.249} & \textbf{24.668} / \textbf{28.438} / \textbf{9.304} & \textbf{23.770} / \textbf{28.200} / \textbf{9.492} \\
\bottomrule
\end{tabular}
}
\end{table*}

\begin{figure}[!t]
	\begin{center}
		\includegraphics[width=\linewidth]{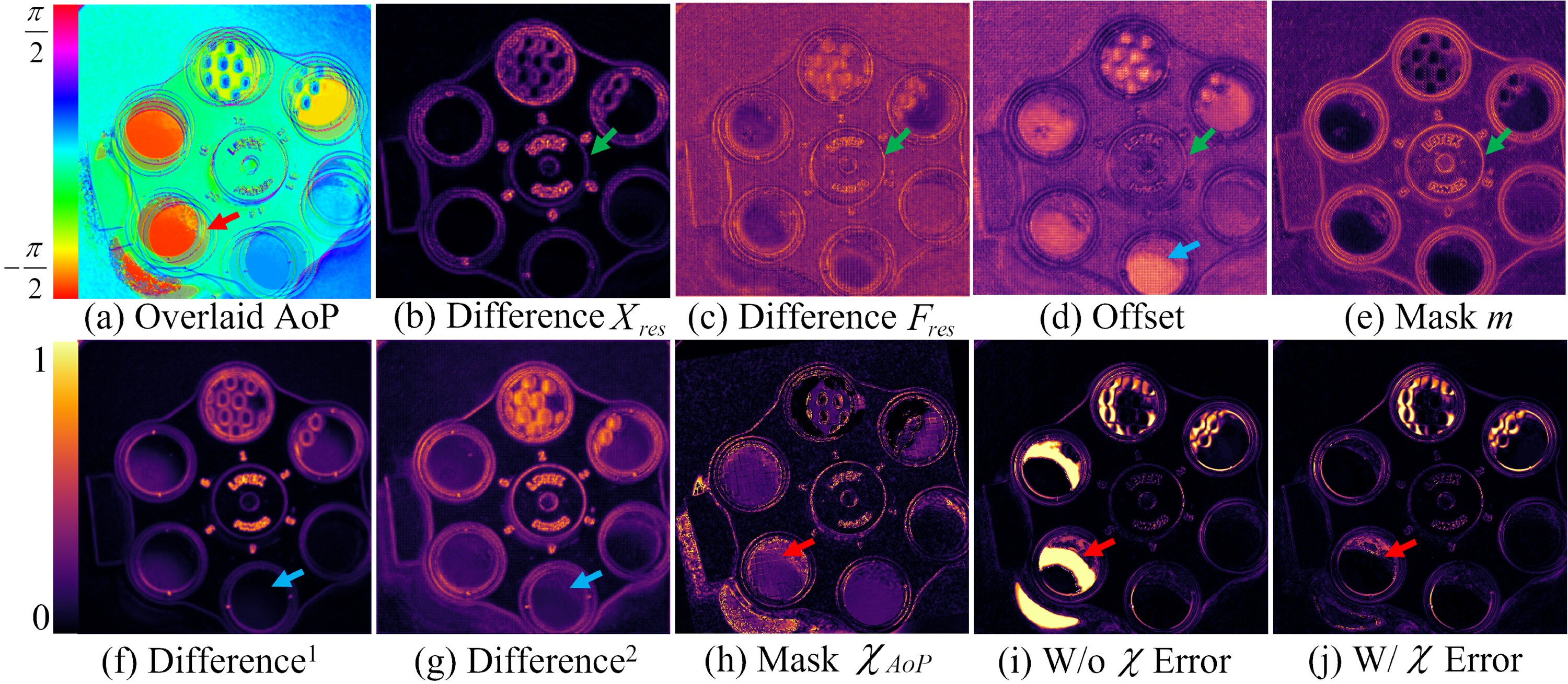}
	\end{center}
	\vspace{-0.8em}
	\caption{Feature visualization. (a) Overlaid AoP at $t=\{0,1\}$. (b)-(c) The differences $X_{res}$ and $F_{res}$ defined in Eq. (\ref{eq:8}). (d)-(e) The DCN offset and mask defined in Eq. (\ref{eq:9}). (f)-(g) Feature differences between $\tilde F_t^{HR}$ and $F_t^{HR}$ in Eq. (\ref{eq:12}), where (f) uses regular convolutional layers and (g) uses the proposed SFT. (h) The mask defined in Eq. (\ref{eq:15}). (i)-(j) Error maps between the reconstructed results and the GT without and with $\chi$.}
	\label{fig:11}
\end{figure}
\textbf{Qualitative evaluation.} Fig. \ref{fig:7} shows visual comparisons for joint demosaicking
($2\times$) and $8\times$ temporal interpolation. Since the VFI module in cascaded approaches only supports $2 \times$ upsampling,
large-motion intermediate frames suffer severe blurring (see frame 5 in Fig. \ref{fig:7}). In contrast,
PIDSR and the proposed method exhibit fewer polarization artifacts than standard three-channel RGB models, which validates the
effectiveness of joint modeling for polarization reconstruction. As shown in Fig. \ref{fig:7}(h), the proposed method achieves
better temporal smoothness than continuous STVSR methods. The unpolarized intensity produced by the proposed method in Fig. \ref{fig:7}(i)
is also sharper than those of competing methods.

Fig. \ref{fig:8} shows $4 \times$ space-time SR. While MoTIF and BF-STVSR suffer from severe polarimetric artifacts, PIDSR and the retrained
BF-STVSR-12ch enhance the accuracy of polarization parameters via unified modeling across polarization channels. In contrast, our method recovers 
the sharpest polarimetric edges with minimal artifacts. Our method also resolves the finest details in intensity images $I$.

Moreover, additional results regarding out-of-distribution spatiotemporal scales are deferred to the supplementary material.

\begin{table}[t]
\centering
\caption{Computational complexity comparison evaluated on $200\times200$ mosaic arrays using an NVIDIA RTX 4090 GPU (demosaicking and $2\times$ frame interpolation).}
\label{tab:4}
\setlength{\tabcolsep}{3.5pt}
\renewcommand{\arraystretch}{1.05}
\begin{tabular}{lccc}
\toprule
Method & Params (M) & FLOPs (T) & Time (ms) \\
\midrule
SCUBA \cite{zhang2025polarization}+ATD \cite{zhang2026atd}       & 17.364+0.753  & 0.403             & 548.92 \\
SCUBA+PIDSR \cite{zhou2025pidsr}     & 17.364+7.647  & \textbf{0.195}    & 397.26 \\
SCUBA+PUGDiff \cite{li2026pugdiff}   & 17.364+899.190 & 26.146            & 452.99 \\
VFIT \cite{lu2022video}+ATD        & 29.054+0.753  & 0.532             & 300.48 \\
VFIT+PIDSR      & 29.054+7.647  & 0.324             & 156.72 \\
VFIT+PUGDiff    & 29.054+899.190 & 26.175            & 267.35 \\
SuperSloMo \cite{jiang2018super}+ATD         & 39.610+0.753  & 0.522             & 280.88 \\
SuperSloMo+PIDSR       & 39.610+7.647  & \underline{0.313} & \underline{138.01} \\
SuperSloMo+PUGDiff     & 39.610+899.190 & 26.170            & 284.90 \\
ZoomingSlowMo \cite{xiang2020zooming}             & \textbf{10.522}& 2.318            & 299.40 \\
TMNet \cite{xu2021temporal}           & 11.618        & 3.343             & 447.71 \\
VideoINR \cite{chen2022videoinr}        & \underline{10.732} & 2.126        & 434.68 \\
MoTIF \cite{chen2023motif}           & 13.195        & 2.227             & 419.24 \\
BF-STVSR \cite{kim2025bf}        & 12.906        & 2.141             & 339.16 \\
\midrule
Ours            & 17.684        & 1.228             & \textbf{103.86} \\
\bottomrule
\end{tabular}
\end{table}
\subsection{Computational Complexity}

Table \ref{tab:4} compares the computational complexity of different paradigms. Cascaded methods have the highest parameter counts and inference times. The proposed method maintains a parameter size similar to standard STVSR methods because it uses comparable network depth and base channel capacity. Its key difference is unified modeling of polarization channels. Jointly processing these channels compresses the effective width allocated to each direction and enables efficient inference. Thus, the proposed method achieves lower FLOPs and shorter inference time than standard STVSR baselines.


\begin{table*}[t]
\centering
\caption{Ablation study on demosaicking ($2\times$) and $8\times$ frame interpolation. ``CLIF'' denotes the basic structure of $f_{pa}$, and ``$\varphi$'' and ``$\mathcal{A}$'' represent the embeddings of polarization and the imaging operator, respectively. ``$\chi$'' indicates the mask used in $\mathcal{L}_{polar}$.}
\label{tab:5}
\renewcommand{\arraystretch}{1.12}

\small 
\setlength{\tabcolsep}{3pt} 

\begin{tabular*}{\textwidth}{@{\extracolsep{\fill}}ccccccccccccc}
\toprule
Config & CLIF & $\varphi$ & $\mathcal{A}$ & MCFR & $\mathcal{L}_{pix}$ & $\mathcal{L}_{polar}$ w/o $\chi$ & $\mathcal{L}_{polar}$ w/ $\chi$ 
& PSNR$_I \uparrow$ & PSNR$_P \uparrow$ & SSIM$_I \uparrow$ & SSIM$_P \uparrow$ & MAE$\downarrow$ \\
\midrule
(a) &        &        &        &\checkmark        & \checkmark &        & \checkmark 
    & 27.830 & 29.021 & 0.855 & 0.753 & 9.278 \\
(b) & \checkmark &        &        &\checkmark        & \checkmark &        & \checkmark 
    & 29.059 & 31.432 & 0.870 & 0.803 & 7.098 \\
(c) & \checkmark & \checkmark &        & \checkmark       & \checkmark &        & \checkmark 
    & 29.098 & \underline{31.459} & \underline{0.870} & \underline{0.805} & \underline{6.912} \\
(d) & \checkmark & \checkmark & \checkmark &        & \checkmark &        & \checkmark 
    & 28.809 & 31.042 & 0.869 & 0.799 & 6.940\\
(e) & \checkmark & \checkmark & \checkmark & \checkmark & \checkmark &        &        
    & \underline{29.245} & 31.041 & 0.870 & 0.797 & 7.777 \\
(f) & \checkmark & \checkmark & \checkmark & \checkmark & \checkmark & \checkmark &        
    & 29.140 & 31.009 & 0.870 & 0.800 & 7.810 \\
\midrule
(g) & \checkmark & \checkmark & \checkmark & \checkmark & \checkmark &        & \checkmark 
    & \textbf{29.331} & \textbf{31.626} & \textbf{0.872} & \textbf{0.809} & \textbf{6.763} \\
\bottomrule
\end{tabular*}
\end{table*}

\subsection{Ablation Study}
We conduct ablations on architectural design, motion field estimation, and loss functions. Tables \ref{tab:5} and \ref{tab:6} and Figs. \ref{fig:10} and \ref{fig:11}
present the performance of different configurations.

\textbf{Architectural design analysis.} The network builds on standard continuous STVSR architectures.
The first modification is unified modeling of all polarization directions, whose effect has been discussed in the baseline comparison.
This unified approach uses inter-direction correlations during feature processing,
thereby suppressing artifacts and producing clean polarization parameter estimation.

We next investigate how the sampling function $f_{pa}$ affects performance.
Table \ref{tab:5}(a)-(b) shows that replacing SIREN with CLIF as the INR improves all metrics, while
Fig. \ref{fig:10}(a) confirms that removing CLIF introduces visible artifacts. These results show the
benefit of using convolutions within the INR. Table \ref{tab:5}(c), (d) and (g) shows that both the polarization
embedding and the forward operator embedding provide consistent metric improvements. As shown in Fig. \ref{fig:10}(b)-(c),
removing these modules causes the light spot in the blue box to have irregular shapes. To further analyze the forward operator
embedding, we visualize its features in Fig. \ref{fig:11}(b)-(e). The residuals $X_{res}$ and $F_{res}$ defined in Eq. (\ref{eq:8})
show high responses at edges, capturing regional information loss from the forward imaging process to guide the DCN. Therefore,
the DCN offset and mask in Fig. \ref{fig:11}(d)-(e) display texture-aware properties. The offset is also strongly
activated at the blue arrow, corresponding to the polarization parameters of the scene.

We also evaluate MCFR. As a feature refinement mechanism before decoding,
MCFR reduces warping-induced errors and yields clear quantitative improvements.
Removing MCFR results in a noticeable performance drop in Table \ref{tab:5}(d). As shown in Fig. \ref{fig:10}(d), removing MCFR introduces ghosting in the red box and scatters the light spot.
Fig. \ref{fig:11}(f)-(g) further compares features when MCFR is replaced with a standard residual block. Compared with
Fig. \ref{fig:11}(f), Fig. \ref{fig:11}(g) shows that the features refined by MCFR change in regions with active polarization responses.
This indicates that MCFR uses polarization and temporal motion cues for targeted refinement, supporting
faithful inter-frame reconstruction.

\textbf{Necessity of optical flow.} Table \ref{tab:6} shows the impact of different motion field estimation
configurations. Removing explicit optical flow and relying on implicit motion field estimation (as in \cite{kim2025bf})
fails to produce an accurate motion field for this task, leading to lower quantitative metrics.
As shown in Fig. \ref{fig:10}(e), the absence of optical flow causes distortions in inter-frame reconstruction.
This may occur because hardware conditions give the polarization benchmark lower temporal consistency than RGB datasets.
Therefore, a pre-trained optical flow estimator is important for a reliable initial motion field. Replacing
the default estimator with an alternative model (e.g., RAFT-lite \cite{teed2020raft}) also reduces performance, 
indicating that the quality of the initial motion estimation affects final network performance. We further investigate
the formulation of $f_{st}$. Replacing SIREN with the architecture from PAINR within $f_{st}$ yields negligible
performance differences, showing that SIREN is sufficient for smooth field estimation. Because PAINR introduces additional
computational overhead, we retain the SIREN-based configuration.

\begin{table}[t]
\centering
\caption{Ablation study of different optical flow estimation configurations on demosaicking ($2\times$) and $8\times$ frame interpolation.}
\label{tab:6}
\setlength{\tabcolsep}{3pt}
\renewcommand{\arraystretch}{1.05}
\resizebox{\columnwidth}{!}{
\begin{tabular}{lccccc}
\toprule
Method & PSNR$_{I}\uparrow$ & PSNR$_{p}\uparrow$ & SSIM$_{I}\uparrow$ & SSIM$_{p}\uparrow$ & MAE$\downarrow$ \\
\midrule
W/o flow       & 26.439 & 29.796 & 0.815 & 0.766 & 7.987 \\
RAFT-lite      & 29.070 & 31.459 & 0.869 & 0.805 & 6.890 \\
PAINR as $f_{st}$& \underline{29.268} & \underline{31.599} & \underline{0.870} & \underline{0.808} & \underline{6.852} \\
\midrule
Ours           & \textbf{29.331} & \textbf{31.626} & \textbf{0.872} & \textbf{0.809} & \textbf{6.763} \\
\bottomrule
\end{tabular}
}
\end{table}

\textbf{Ablation on loss functions.} We ablate the loss components.
${{\cal L}_{pix}}$ constrains the four directions against the GT to ensure basic intensity reconstruction.
However, polarization recovery requires specific polarimetric constraints. As shown in Table \ref{tab:5}(e)-(f),
direct polarization constraints fail to improve the metrics, likely because the abstract polarization
domain and computational noise disrupt the balance between intensity and polarization optimization.
In contrast, the flow-guided polarization variation mask forces the network to focus on regions with
active polarization changes (Fig. \ref{fig:11}(h)), while ${{\cal L}_{sm}}$ ensures smoothness elsewhere.
By constraining the optimization space, this mechanism helps the network capture polarization
dynamics instead of overfitting to flat backgrounds or noise. Consequently, Fig. \ref{fig:10}(f)-(h) shows
that the proposed loss gives better visual quality than alternative configurations. Moreover,
Fig. \ref{fig:11}(i)-(j) shows that the proposed method better estimates inter-frame polarization variations,
and Table \ref{tab:5}(g) shows that the full configuration achieves the best performance.

\section{Conclusion}
This paper presents the first framework for space-time polarization video reconstruction. Built on a continuous STVSR architecture, the method introduces designs customized for polarization imaging. It uniformly models polarization directions and uses unpolarized intensity for motion field estimation. Polarization and imaging operator embeddings improve the sampling function for polarization reconstruction, and motion cues reduce feature warping errors. A flow-guided polarization variation loss further constrains the network to learn polarization evolution patterns. To support this research direction, we also present a large-scale color polarization video benchmark. Extensive experiments, including demosaicking and super-resolution results, demonstrate the effectiveness of the proposed method.

\section*{Acknowledgments}
This work was supported in part by the National Natural Science Foundation of China (Grant No. 62105372), the National Natural Science Foundation of China (Grant
No. 62471497), the Fundamental Research Foundation of National Key Laboratory of Automatic Target Recognition (Grant Number: WDZC20255290209), and the Hunan Provincial Research and Development Project (Grant No.2025QK3019).
This work was supported in part by the High Performance Computing Center of Central South University.

\bibliographystyle{IEEEtran}
\bibliography{ref}

@article{luo2025cpifuse,
  title={CPIFuse: Toward realistic color and enhanced textures in color polarization image fusion},
  author={Luo, Yidong and Zhang, Junchao and Li, Chenggong},
  journal={Information Fusion},
  volume={120},
  pages={103111},
  year={2025},
  publisher={Elsevier}
}

@inproceedings{rahman2025polarization,
  title={Polarization Denoising and Demosaicking: Dataset and Baseline Method},
  author={Rahman, Muhamad Daniel Ariff Bin Abdul and Monno, Yusuke and Tanaka, Masayuki and Okutomi, Masatoshi},
  booktitle={IEEE International Conference on Image Processing},
  pages={2724--2729},
  year={2025},
  organization={IEEE}
}

@article{zhu2024podb,
  title={PODB: A learning-based polarimetric object detection benchmark for road scenes in adverse weather conditions},
  author={Zhu, Zhen and Li, Xiaobo and Zhai, Jingsheng and Hu, Haofeng},
  journal={Information Fusion},
  volume={108},
  pages={102385},
  year={2024},
  publisher={Elsevier}
}

@inproceedings{jeon2024spectral,
  title={Spectral and polarization vision: Spectro-polarimetric real-world dataset},
  author={Jeon, Yujin and Choi, Eunsue and Kim, Youngchan and Moon, Yunseong and Omer, Khalid and Heide, Felix and Baek, Seung-Hwan},
  booktitle={Proceedings of the IEEE/CVF Conference on Computer Vision and Pattern Recognition},
  pages={22098--22108},
  year={2024}
}

@inproceedings{liu2025slam3r,
  title={Slam3r: Real-time dense scene reconstruction from monocular rgb videos},
  author={Liu, Yuzheng and Dong, Siyan and Wang, Shuzhe and Yin, Yingda and Yang, Yanchao and Fan, Qingnan and Chen, Baoquan},
  booktitle={Proceedings of the Computer Vision and Pattern Recognition Conference},
  pages={16651--16662},
  year={2025}
}

@inproceedings{hong2025motionbench,
  title={Motionbench: Benchmarking and improving fine-grained video motion understanding for vision language models},
  author={Hong, Wenyi and Cheng, Yean and Yang, Zhuoyi and Wang, Weihan and Wang, Lefan and Gu, Xiaotao and Huang, Shiyu and Dong, Yuxiao and Tang, Jie},
  booktitle={Proceedings of the Computer Vision and Pattern Recognition Conference},
  pages={8450--8460},
  year={2025}
}

@article{wu2021polarization,
  title={Polarization image demosaicking using polarization channel difference prior},
  author={Wu, Rongyuan and Zhao, Yongqiang and Li, Ning and Kong, Seong G},
  journal={Optics Express},
  volume={29},
  number={14},
  pages={22066--22079},
  year={2021},
  publisher={Optical Society of America}
}

@article{ponimatkin20256d,
  title={6d object pose tracking in internet videos for robotic manipulation},
  author={Ponimatkin, Georgy and C{\'\i}fka, Martin and Sou{\v{c}}ek, Tom{\'a}{\v{s}} and Fourmy, M{\'e}d{\'e}ric and Labb{\'e}, Yann and Petrik, Vladimir and Sivic, Josef},
  journal={arXiv preprint arXiv:2503.10307},
  year={2025}
}

@inproceedings{chen2025rethinking,
  title={Rethinking temporal fusion with a unified gradient descent view for 3d semantic occupancy prediction},
  author={Chen, Dubing and Zheng, Huan and Fang, Jin and Dong, Xingping and Li, Xianfei and Liao, Wenlong and He, Tao and Peng, Pai and Shen, Jianbing},
  booktitle={Proceedings of the IEEE/CVF Conference on Computer Vision and Pattern Recognition},
  pages={1505--1515},
  year={2025}
}

@article{zhang2017pca,
  title={PCA-based denoising method for division of focal plane polarimeters},
  author={Zhang, Junchao and Luo, Haibo and Liang, Rongguang and Zhou, Wei and Hui, Bin and Chang, Zheng},
  journal={Optics Express},
  volume={25},
  number={3},
  pages={2391--2400},
  year={2017},
  publisher={Optical Society of America}
}

@article{koenderink1984structure,
  title={The structure of images},
  author={Koenderink, Jan J},
  journal={Biological Cybernetics},
  volume={50},
  number={5},
  pages={363--370},
  year={1984},
  publisher={Springer}
}

@article{chung2022diffusion,
  title={Diffusion posterior sampling for general noisy inverse problems},
  author={Chung, Hyungjin and Kim, Jeongsol and Mccann, Michael T and Klasky, Marc L and Ye, Jong Chul},
  journal={arXiv preprint arXiv:2209.14687},
  year={2022}
}

@article{chihaoui2024blind,
  title={Blind image restoration via fast diffusion inversion},
  author={Chihaoui, Hamadi and Lemkhenter, Abdelhak and Favaro, Paolo},
  journal={Advances in Neural Information Processing Systems},
  volume={37},
  pages={34513--34532},
  year={2024}
}

@article{wang2022zero,
  title={Zero-shot image restoration using denoising diffusion null-space model},
  author={Wang, Yinhuai and Yu, Jiwen and Zhang, Jian},
  journal={arXiv preprint arXiv:2212.00490},
  year={2022}
}

@article{li2019demosaicking,
  title={Demosaicking DoFP images using Newton’s polynomial interpolation and polarization difference model},
  author={Li, Ning and Zhao, Yongqiang and Pan, Quan and Kong, Seong G},
  journal={Optics Express},
  volume={27},
  number={2},
  pages={1376--1391},
  year={2019},
  publisher={Optical Society of America}
}

@inproceedings{li2026pugdiff,
  title={Polarization Uncertainty-Guided Diffusion Model for Color Polarization Image Demosaicking},
  author={Li, Chenggong and Luo, Yidong and Zhang, Junchao and Yang Degui},
  booktitle={Proceedings of the AAAI Conference on Artificial Intelligence},
  number={8},
  volume={40},
  pages={6028-6036},
  year={2026}
  }

@article{zhang2026atd,
  title={ATD: Improved Transformer With Adaptive Token Dictionary for Image Restoration},
  author={Zhang, Leheng and Long, Wei and Li, Yawei and Zhou, Xingyu and Zhao, Xiaorui and Gu, Shuhang},
  journal={IEEE Transactions on Pattern Analysis and Machine Intelligence},
  year={2026},
  publisher={IEEE}
}

@article{yu2023color,
  title={Color polarization image super-resolution reconstruction via a cross-branch supervised learning strategy},
  author={Yu, Dabing and Li, Qingwu and Zhang, Zhiliang and Huo, Guanying and Xu, Chang and Zhou, Yaqin},
  journal={Optics and Lasers in Engineering},
  volume={165},
  pages={107469},
  year={2023},
  publisher={Elsevier}
}

@article{hu2023polarized,
  title={Polarized image super-resolution via a deep convolutional neural network},
  author={Hu, Haofeng and Yang, Shiyao and Li, Xiaobo and Cheng, Zhenzhou and Liu, Tiegen and Zhai, Jingsheng},
  journal={Optics Express},
  volume={31},
  number={5},
  pages={8535--8547},
  year={2023},
  publisher={Optica Publishing Group}
}

@inproceedings{deschaintre2021deep,
  title={Deep polarization imaging for 3d shape and svbrdf acquisition},
  author={Deschaintre, Valentin and Lin, Yiming and Ghosh, Abhijeet},
  booktitle={Proceedings of the IEEE/CVF Conference on Computer Vision and Pattern Recognition},
  pages={15567--15576},
  year={2021}
}

@article{lyu2023shape,
  title={Shape from polarization with distant lighting estimation},
  author={Lyu, Youwei and Zhao, Lingran and Li, Si and Shi, Boxin},
  journal={IEEE Transactions on Pattern Analysis and Machine Intelligence},
  volume={45},
  number={11},
  pages={13991--14004},
  year={2023},
  publisher={IEEE}
}

@article{guo2024attention,
  title={Attention-based progressive discrimination generative adversarial networks for polarimetric image demosaicing},
  author={Guo, Yuxuan and Dai, Xiaobing and Wang, Shaoju and Jin, Guang and Zhang, Xuemin},
  journal={IEEE Transactions on Computational Imaging},
  year={2024},
  volume={10},
  number={},
  pages={713-725},
  publisher={IEEE}
}

@inproceedings{qiu2021linear,
  title={Linear polarization demosaicking for monochrome and colour polarization focal plane arrays},
  author={Qiu, Simeng and Fu, Qiang and Wang, Congli and Heidrich, Wolfgang},
  booktitle={Computer Graphics Forum},
  volume={40},
  number={6},
  pages={77--89},
  year={2021},
  organization={Wiley Online Library}
}

@article{wen2019convolutional,
  title={Convolutional demosaicing network for joint chromatic and polarimetric imagery},
  author={Wen, Sijia and Zheng, Yinqiang and Lu, Feng and Zhao, Qinping},
  journal={Optics Letters},
  volume={44},
  number={22},
  pages={5646--5649},
  year={2019},
  publisher={Optical Society of America}
}

@inproceedings{mei2023deep,
  title={Deep polarization reconstruction with PDAVIS events},
  author={Mei, Haiyang and Wang, Zuowen and Yang, Xin and Wei, Xiaopeng and Delbruck, Tobi},
  booktitle={Proceedings of the IEEE/CVF Conference on Computer Vision and Pattern Recognition},
  pages={22149--22158},
  year={2023}
}

@article{li2025demosaicking,
  title={Demosaicking customized diffusion model for snapshot polarization imaging},
  author={Li, Chenggong and Luo, Yidong and Wu, Caiyun and Zhang, Junchao and Yang, Degui and Zhao, Dangjun},
  journal={Optics \& Laser Technology},
  volume={188},
  pages={112868},
  year={2025},
  publisher={Elsevier}
}

@article{luo2023sparse,
  title={Sparse representation-based demosaicking method for joint chromatic and polarimetric imagery},
  author={Luo, Yidong and Zhang, Junchao and Tian, Di},
  journal={Optics and Lasers in Engineering},
  volume={164},
  pages={107526},
  year={2023},
  publisher={Elsevier}
}

@inproceedings{nguyen2022two,
  title={Two-step color-polarization demosaicking network},
  author={Nguyen, Vy and Tanaka, Masayuki and Monno, Yusuke and Okutomi, Masatoshi},
  booktitle={IEEE International Conference on Image Processing},
  pages={1011--1015},
  year={2022},
}

@article{zhang2018learning,
  title={Learning a convolutional demosaicing network for microgrid polarimeter imagery},
  author={Zhang, Junchao and Shao, Jianbo and Luo, Haibo and Zhang, Xiangyue and Hui, Bin and Chang, Zheng and Liang, Rongguang},
  journal={Optics Letters},
  volume={43},
  number={18},
  pages={4534--4537},
  year={2018},
  publisher={Optical Society of America}
}

@article{sun2021color,
  title={Color polarization demosaicking by a convolutional neural network},
  author={Sun, Yuanyuan and Zhang, Junchao and Liang, Rongguang},
  journal={Optics Letters},
  volume={46},
  number={17},
  pages={4338--4341},
  year={2021},
  publisher={Optical Society of America}
}

@article{luo2024learning,
  title={Learning a non-locally regularized convolutional sparse representation for joint chromatic and polarimetric demosaicking},
  author={Luo, Yidong and Zhang, Junchao and Shao, Jianbo and Tian, Jiandong and Ma, Jiayi},
  journal={IEEE Transactions on Image Processing},
  year={2024},
  volume={33},
  number={},
  pages={5029-5044},
  publisher={IEEE}
}

@article{wen2021sparse,
  title={A sparse representation based joint demosaicing method for single-chip polarized color sensor},
  author={Wen, Sijia and Zheng, Yinqiang and Lu, Feng},
  journal={IEEE Transactions on Image Processing},
  volume={30},
  pages={4171--4182},
  year={2021},
  publisher={IEEE}
}

@inproceedings{haris2020space,
  title={Space-time-aware multi-resolution video enhancement},
  author={Haris, Muhammad and Shakhnarovich, Greg and Ukita, Norimichi},
  booktitle={Proceedings of the IEEE/CVF Conference on Computer Vision and Pattern Recognition},
  pages={2859--2868},
  year={2020}
}

@inproceedings{xiang2020zooming,
  title={Zooming slow-mo: Fast and accurate one-stage space-time video super-resolution},
  author={Xiang, Xiaoyu and Tian, Yapeng and Zhang, Yulun and Fu, Yun and Allebach, Jan P and Xu, Chenliang},
  booktitle={Proceedings of the IEEE/CVF Conference on Computer Vision and Pattern Recognition},
  pages={3370--3379},
  year={2020}
}

@inproceedings{xu2021temporal,
  title={Temporal modulation network for controllable space-time video super-resolution},
  author={Xu, Gang and Xu, Jun and Li, Zhen and Wang, Liang and Sun, Xing and Cheng, Ming-Ming},
  booktitle={Proceedings of the IEEE/CVF Conference on Computer Vision and Pattern Recognition},
  pages={6388--6397},
  year={2021}
}

@article{wei2026osdenhancer,
  title={OSDEnhancer: Taming Real-World Space-Time Video Super-Resolution with One-Step Diffusion},
  author={Wei, Shuoyan and Li, Feng and Zhou, Chen and Cong, Runmin and Zhao, Yao and Bai, Huihui},
  journal={arXiv preprint arXiv:2601.20308},
  year={2026}
}

@article{he2024venhancer,
  title={Venhancer: Generative space-time enhancement for video generation},
  author={He, Jingwen and Xue, Tianfan and Liu, Dongyang and Lin, Xinqi and Gao, Peng and Lin, Dahua and Qiao, Yu and Ouyang, Wanli and Liu, Ziwei},
  journal={arXiv preprint arXiv:2407.07667},
  year={2024}
}

@inproceedings{morimatsu2020monochrome,
  title={Monochrome and color polarization demosaicking using edge-aware residual interpolation},
  author={Morimatsu, Miki and Monno, Yusuke and Tanaka, Masayuki and Okutomi, Masatoshi},
  booktitle={IEEE International Conference on Image Processing},
  pages={2571--2575},
  year={2020},
  organization={IEEE}
}

@inproceedings{hwang2025benchmarking,
  title={Benchmarking Burst Super-Resolution for Polarization Images: Noise Dataset and Analysis},
  author={Hwang, Inseung and Choi, Kiseok and Ha, Hyunho and Kim, Min H},
  booktitle={Proceedings of the IEEE/CVF International Conference on Computer Vision},
  pages={24899--24909},
  year={2025}
}

@article{lyu2022physics,
  title={Physics-guided reflection separation from a pair of unpolarized and polarized images},
  author={Lyu, Youwei and Cui, Zhaopeng and Li, Si and Pollefeys, Marc and Shi, Boxin},
  journal={IEEE Transactions on Pattern Analysis and Machine Intelligence},
  volume={45},
  number={2},
  pages={2151--2165},
  year={2022},
  publisher={IEEE}
}

@inproceedings{zhou2025pidsr,
  title={PIDSR: Complementary Polarized Image Demosaicing and Super-Resolution},
  author={Zhou, Shuangfan and Zhou, Chu and Lyu, Youwei and Guo, Heng and Ma, Zhanyu and Shi, Boxin and Sato, Imari},
  booktitle={Proceedings of the Computer Vision and Pattern Recognition Conference},
  pages={16081--16090},
  year={2025}
}

@article{jaderberg2015spatial,
  title={Spatial transformer networks},
  author={Jaderberg, Max and Simonyan, Karen and Zisserman, Andrew and others},
  journal={Advances in Neural Information Processing Systems},
  volume={28},
  year={2015}
}

@article{liu2025sharecmp,
  title={Sharecmp: Polarization-aware rgb-p semantic segmentation},
  author={Liu, Zhuoyan and Wang, Bo and Wang, Lizhi and Mao, Chenyu and Li, Ye},
  journal={IEEE Transactions on Circuits and Systems for Video Technology},
  year={2025},
  publisher={IEEE}
}

@inproceedings{yao2025polarfree,
  title={PolarFree: Polarization-based Reflection-Free Imaging},
  author={Yao, Mingde and Wang, Menglu and Tam, King-Man and Li, Lingen and Xue, Tianfan and Gu, Jinwei},
  booktitle={Proceedings of the Computer Vision and Pattern Recognition Conference},
  pages={10890--10899},
  year={2025}
}

@inproceedings{lei2020polarized,
  title={Polarized reflection removal with perfect alignment in the wild},
  author={Lei, Chenyang and Huang, Xuhua and Zhang, Mengdi and Yan, Qiong and Sun, Wenxiu and Chen, Qifeng},
  booktitle={Proceedings of the IEEE/CVF Conference on Computer Vision and Pattern Recognition},
  pages={1750--1758},
  year={2020}
}

@inproceedings{lei2022shape,
  title={Shape from polarization for complex scenes in the wild},
  author={Lei, Chenyang and Qi, Chenyang and Xie, Jiaxin and Fan, Na and Koltun, Vladlen and Chen, Qifeng},
  booktitle={Proceedings of the IEEE/CVF Conference on Computer Vision and Pattern Recognition},
  pages={12632--12641},
  year={2022}
}

@inproceedings{jiang2018super,
  title={Super slomo: High quality estimation of multiple intermediate frames for video interpolation},
  author={Jiang, Huaizu and Sun, Deqing and Jampani, Varun and Yang, Ming-Hsuan and Learned-Miller, Erik and Kautz, Jan},
  booktitle={Proceedings of the IEEE/CVF Conference on Computer Vision and Pattern Recognition},
  pages={9000--9008},
  year={2018}
}

@inproceedings{lu2022video,
  title={Video frame interpolation with transformer},
  author={Lu, Liying and Wu, Ruizheng and Lin, Huaijia and Lu, Jiangbo and Jia, Jiaya},
  booktitle={Proceedings of the IEEE/CVF Conference on Computer Vision and Pattern Recognition},
  pages={3532--3542},
  year={2022}
}

@article{chen2025invertible,
  title={Invertible diffusion models for compressed sensing},
  author={Chen, Bin and Zhang, Zhenyu and Li, Weiqi and Zhao, Chen and Yu, Jiwen and Zhao, Shijie and Chen, Jie and Zhang, Jian},
  journal={IEEE Transactions on Pattern Analysis and Machine Intelligence},
  volume={47},
  number={5},
  pages={3992--4006},
  year={2025},
  publisher={IEEE}
}

@article{saharia2022image,
  title={Image super-resolution via iterative refinement},
  author={Saharia, Chitwan and Ho, Jonathan and Chan, William and Salimans, Tim and Fleet, David J and Norouzi, Mohammad},
  journal={IEEE Transactions on Pattern Analysis and Machine Intelligence},
  volume={45},
  number={4},
  pages={4713--4726},
  year={2022},
  publisher={IEEE}
}

@inproceedings{chen2021learning,
  title={Learning continuous image representation with local implicit image function},
  author={Chen, Yinbo and Liu, Sifei and Wang, Xiaolong},
  booktitle={Proceedings of the IEEE/CVF Conference on Computer Vision and Pattern Recognition},
  pages={8628--8638},
  year={2021}
}

@inproceedings{tzabari2020polarized,
  title={Polarized optical-flow gyroscope},
  author={Tzabari, Masada and Schechner, Yoav Y},
  booktitle={Proceedings of the European Conference on Computer Vision},
  pages={363--381},
  year={2020},
}

@article{xu2019quadratic,
  title={Quadratic video interpolation},
  author={Xu, Xiangyu and Siyao, Li and Sun, Wenxiu and Yin, Qian and Yang, Ming-Hsuan},
  journal={Advances in Neural Information Processing Systems},
  volume={32},
  year={2019}
}

@inproceedings{chan2022basicvsr,
  title={Basicvsr++: Improving video super-resolution with enhanced propagation and alignment},
  author={Chan, Kelvin CK and Zhou, Shangchen and Xu, Xiangyu and Loy, Chen Change},
  booktitle={Proceedings of the IEEE/CVF Conference on Computer Vision and Pattern Recognition},
  pages={5972--5981},
  year={2022}
}

@inproceedings{rebhan2019principle,
  title={Principle investigations on polarization image sensors},
  author={Rebhan, David and Rosenberger, Maik and Notni, Gunther},
  booktitle={Photonics and Education in Measurement Science 2019},
  volume={11144},
  pages={50--54},
  year={2019},
  organization={SPIE}
}

@book{collett2005field,
  title={Field guide to polarization},
  author={Collett, Edward},
  volume={15},
  year={2005},
  publisher={SPIE press Bellingham}
}

@article{zhang2025polarization,
  title={Polarization video frame interpolation for 3D human pose reconstruction with attention mechanism},
  author={Zhang, Xin and Wang, Xuesong and Xu, Yixuan and Wu, Xianyu and Huang, Feng},
  journal={Optics and Lasers in Engineering},
  volume={193},
  pages={109046},
  year={2025},
  publisher={Elsevier}
}

@inproceedings{niklaus2020softmax,
  title={Softmax splatting for video frame interpolation},
  author={Niklaus, Simon and Liu, Feng},
  booktitle={Proceedings of the IEEE/CVF Conference on Computer Vision and Pattern Recognition},
  pages={5437--5446},
  year={2020}
}

@inproceedings{zhu2019deformable,
  title={Deformable convnets v2: More deformable, better results},
  author={Zhu, Xizhou and Hu, Han and Lin, Stephen and Dai, Jifeng},
  booktitle={Proceedings of the IEEE/CVF Conference on Computer Vision and Pattern Recognition},
  pages={9308--9316},
  year={2019}
}

@article{he2012guided,
  title={Guided image filtering},
  author={He, Kaiming and Sun, Jian and Tang, Xiaoou},
  journal={IEEE transactions on pattern analysis and machine intelligence},
  volume={35},
  number={6},
  pages={1397--1409},
  year={2012},
  publisher={IEEE}
}

@inproceedings{lai2018learning,
  title={Learning blind video temporal consistency},
  author={Lai, Wei-Sheng and Huang, Jia-Bin and Wang, Oliver and Shechtman, Eli and Yumer, Ersin and Yang, Ming-Hsuan},
  booktitle={Proceedings of the European Conference on Computer Vision},
  pages={170--185},
  year={2018}
}

@inproceedings{chen2024image,
  title={Image neural field diffusion models},
  author={Chen, Yinbo and Wang, Oliver and Zhang, Richard and Shechtman, Eli and Wang, Xiaolong and Gharbi, Michael},
  booktitle={Proceedings of the IEEE/CVF Conference on Computer Vision and Pattern Recognition},
  pages={8007--8017},
  year={2024}
}

@inproceedings{wang2018recovering,
  title={Recovering realistic texture in image super-resolution by deep spatial feature transform},
  author={Wang, Xintao and Yu, Ke and Dong, Chao and Loy, Chen Change},
  booktitle={Proceedings of the IEEE/CVF Conference on Computer Vision and Pattern Recognition},
  pages={606--615},
  year={2018}
}

@article{sitzmann2020implicit,
  title={Implicit neural representations with periodic activation functions},
  author={Sitzmann, Vincent and Martel, Julien and Bergman, Alexander and Lindell, David and Wetzstein, Gordon},
  journal={Advances in Neural Information Processing Systems},
  volume={33},
  pages={7462--7473},
  year={2020}
}

@inproceedings{wang2024sea,
  title={Sea-raft: Simple, efficient, accurate raft for optical flow},
  author={Wang, Yihan and Lipson, Lahav and Deng, Jia},
  booktitle={Proceedings of the European Conference on Computer Vision},
  pages={36--54},
  year={2024},
}

@inproceedings{chen2022videoinr,
  title={Videoinr: Learning video implicit neural representation for continuous space-time super-resolution},
  author={Chen, Zeyuan and Chen, Yinbo and Liu, Jingwen and Xu, Xingqian and Goel, Vidit and Wang, Zhangyang and Shi, Humphrey and Wang, Xiaolong},
  booktitle={Proceedings of the IEEE/CVF Conference on Computer Vision and Pattern Recognition},
  pages={2047--2057},
  year={2022}
}

@inproceedings{teed2020raft,
  title={Raft: Recurrent all-pairs field transforms for optical flow},
  author={Teed, Zachary and Deng, Jia},
  booktitle={Proceedings of the European Conference on Computer Vision},
  pages={402--419},
  year={2020},
}

@inproceedings{chen2023motif,
  title={MoTIF: Learning motion trajectories with local implicit neural functions for continuous space-time video super-resolution},
  author={Chen, Yi-Hsin and Chen, Si-Cun and Lin, Yen-Yu and Peng, Wen-Hsiao},
  booktitle={Proceedings of the IEEE/CVF International Conference on Computer Vision},
  pages={23131--23141},
  year={2023}
}

@inproceedings{kim2025bf,
  title={BF-STVSR: B-Splines and Fourier---Best Friends for High Fidelity Spatial-Temporal Video Super-Resolution},
  author={Kim, Eunjin and Kim, Hyeonjin and Jin, Kyong Hwan and Yoo, Jaejun},
  booktitle={Proceedings of the IEEE/CVF Conference on Computer Vision and Pattern Recognition},
  pages={28009--28018},
  year={2025}
}

@article{becker2025continuous,
  title={Continuous Space-Time Video Super-Resolution with 3D Fourier Fields},
  author={Becker, Alexander and Erbach, Julius and Narnhofer, Dominik and Schindler, Konrad},
  journal={arXiv preprint arXiv:2509.26325},
  year={2025}
}
\end{document}